\documentclass[10pt,a4paper]{article}

\usepackage[final]{lrec2026} 

\usepackage{graphicx}

\usepackage{microtype}
\usepackage{amsmath}
\usepackage{tabularx}

\usepackage{comment}
\usepackage{rotating}
\usepackage{placeins}

\title{Chain-of-Thought Reasoning Improves Context-Aware Translation with Large Language Models} 

\name{Shabnam Ataee$^1$, Hugo Huart$^1$, Andrei Popescu-Belis$^{1,2}$}  

\address{$^1$HEIG-VD/HES-SO, Route de Cheseaux 1, 1401 Yverdon-les-Bains, Switzerland\\
         $^2$EPFL, 1018 Lausanne, Switzerland\\
         \{shabnam.ataee, hugo.huart, andrei.popescu-belis\}@heig-vd.ch\\}

\abstract{
This paper assesses the ability of large language models (LLMs) to translate texts that include inter-sentential dependencies.  We use the English-French DiscEvalMT benchmark \citep{bawden-etal-2018-evaluating} with pairs of sentences containing translation challenges for pronominal anaphora and lexical cohesion.  We evaluate 12 LLMs from the DeepSeek-R1, GPT, Llama, Mistral and Phi families on two tasks: (1)~distinguish a correct translation from a wrong but plausible one; and (2)~generate a correct translation.  We compare prompts that encourage chain-of-thought reasoning with those that do not.  The best models take advantage of reasoning and reach about 90\% accuracy on the first task and COMET scores of about 92\% on the second task, with GPT-4, GPT-4o and Phi standing out.  Moreover, we observe a ``wise get wiser'' effect: the improvements through reasoning are larger for models that already perform well without reasoning.
 \\ \newline  }

\begin{document}

\maketitleabstract

\section{Introduction} 

Large language models (LLMs) have shown impressive capacities for translation, among many other tasks, but their translations are still not perfect \citep{kocmi-etal-2024-findings,cui-etal-2025-multilingual}.  To increase quality, we examine whether chain-of-thought (CoT) reasoning helps to improve referential and lexical cohesion across sentences.  Indeed, when prompted to generate a step-by-step explanation of their answers, LLMs improve substantially on many tasks \citep{wei2022chain}. 

We hypothesize that CoT can improve the translation of elements that maintain coherence across sentences by making explicit translation decisions.  To test our hypothesis, we use the DiscEvalMT English-French test suites for pronominal anaphora and lexical consistency designed by \citet{bawden-etal-2018-evaluating}, in two settings. First, we ask an LLM to select the correct translation among the two contrastive alternatives present in the data.  Second, we ask the LLM to translate each test sentence and we score the result against a reference translation.  In both settings, we demonstrate the advantages of CoT prompting, though only for the largest LLMs, which improve their performance when encouraged to reason before translating. 

The paper is organized as follows.  We review related work in Section~\ref{sec:related-work}.  We present the benchmark data and the evaluation metrics in Section~\ref{sec:data-metrics}.  The tested LLMs are listed in Section~\ref{sec:llms}: they include 4 GPT proprietary models from OpenAI and 8 open-weight models.  
The prompts and the results for the contrastive evaluation task (i.e., selecting the correct translation) are presented in Section~\ref{sec:contrastive-task}, while those for the translation task are in Section~\ref{sec:translation-task}.
Our main contributions are the following ones:
\begin{itemize} \setlength{\itemsep}{0pt}
    \item We assess the capacity of 12 LLMs to translate coherently, showing that they reach a new state of the art on the DiscEvalMT benchmark for pronominal anaphora and lexical cohesion.
    \item We show that the translation task is reliably scored, as BLEU, chrF, BERTscore, and COMET scores are correlated.
    \item We demonstrate that CoT prompting improves coherence, though only for the best models.  In other words, powerful models have a better capacity to leverage reasoning, which we call a ``wise get wiser'' effect.
    \item We share the outputs of the tested systems at \url{https://zenodo.org/records/17483104}.
\end{itemize}

\section{Related Work}
\label{sec:related-work}

The translation capabilities of multilingual LLMs have attracted growing attention since 2023.  Conversational LLMs are now often used for translation tasks, which represent around 4.5\% of non-professional interactions with ChatGPT \citep{chatterji2025howpeople}. Therefore, studies evaluating the translation capability of LLMs are of high importance, all the more that translation is included in common benchmarks for LLMs.  LLMs have been tested since 2023 at the Workshops on Machine Translation \citep{kocmi-EtAl:2023:WMT,kocmi-etal-2024-findings}, and in 2025, the majority of submissions to the general task were LLM-based \citep{kocmi-EtAl:2025:WMT1}.  At the same time, the first studies of LLMs for MT included quantitative evaluations \citep{vilar-etal-2023-prompting,pmlr-v202-zhang23m,hendy2023good}.  Some studies focused on evaluation only: for instance, \citet{bawden-yvon-2023-investigating} demonstrated the strong MT performance of BLOOM, particularly in few-shot settings and for high-resource language pairs.  Fine-tuning of LLMs for document-level MT was found to be moderately beneficial \citep{wu2024adapting}.

Word-sense disambiguation (WSD) for translation is one of the challenges on which LLMs have been evaluated.  Test suites related to WSD include DiscEvalMT \citep{bawden-etal-2018-evaluating} which we use in this paper and describe below, ContraWSD \citep{rios-gonzales-etal-2017-improving} with about 7,000 sentences with reference translations, and a smaller test suite to score word translations \citep{rios-etal-2018-word}.  The DiBiMT benchmark with about 600 examples, for translation from English to five other languages, was proposed by \citet{campolungo-etal-2022-dibimt} but remains private and systems must be submitted to its owners. DiBiMT was used by \citet{iyer-chen-birch:2023:WMT} to score the capacity of LLMs for WSD.

The capacity to translate potentially ambiguous pronouns can be tested using contrastive test suites such as DiscEvalMT \citep{bawden-etal-2018-evaluating}, ContraPRO \citep{muller-etal-2018-large,lopes-etal-2020-document}, or PROTEST \citep{guillou-hardmeier-2016-protest}, which was used at WMT 2018 \citep{guillou-etal-2018-pronoun}. A test suite for ellipses and lexical cohesion was designed by \citet{voita-etal-2019-good}.  A review of scores reached on the DiscEvalMT data by various context-aware MT systems appears below in Section~\ref{sec:contrastive-results}.  An alternative approach is to align and compare pronouns in the MT output with those in a reference translation \cite{miculicich-2017-validation}, but scores may be biased by the choice of the antecedent.

The correct translation of pronouns or ambiguous words is often related to the more general ability to leverage inter-sentential dependencies and generate a contextually-correct translation, reviewed in several studies \citep{popescu2019context,maruf2022survey,jin-etal-2023-challenges,Castilho_Knowles_2025}.  Document-level translation and its evaluation are important because MT outputs may appear competitive with human translations at the sentence level, but not at the document level \citep{laubli-etal-2018-machine}.  An evaluation of GPT-3.5 and GPT-4 on document-level translation \citep{wang2023document} found that they surpassed non-LLM systems in human ratings. \citet{karpinska-iyyer:2023:WMT} found that LLMs translate better entire paragraphs than individual sentences, in the case of literary translation.  In an analysis of the translation of six types of contextual dependencies (with one sentence per type), \citet{castilho-etal-2023-online} found that GPT 3.5 outperformed NMT systems on three high-resource languages, but not on a low-resource one (Gaelic).  Beyond direct translation, interaction was a key element in the WSD process proposed by \citet{pilault-etal-2023-interactive}, using questions and answers as a chain-of-thought input to an LLM.  

Recent work has explored how reasoning can improve LLM-based translation, inspired by the seminal work of \citet{wei2022chain}, who introduced chain-of-thought (CoT) prompting and showed that guiding models through intermediate steps improves performance on complex tasks. \citet{liu2025new} argue that reasoning enhances translation by improving coherence, cultural alignment, and self-reflection. \citet{he2025r1t1} present R1-T1, which uses expert CoT templates and reinforcement learning to encourage inference-time reasoning for translation. \citet{ye2025well} found that reasoning models outperform standard LLMs on semantically complex and domain-specific MT, especially for longer texts. 

From the variety of MT evaluation metrics, many have also been considered for extension to document-level evaluation \citep{dahan2024}.  Given that we only consider here sentence pairs, we will consider four widely used metrics.  We will use BLEU and chrF \citep{papineni-etal-2002-bleu,popovic-2015-chrf}, which are based on surface-level overlap, although they offer limited insights into discourse-level quality.  We will also use embedding-based metrics such as BERTScore \citep{zhang2019bertscore}, and trained metrics such as COMET \citep{rei-etal-2020-comet}, as they model semantic similarity and are now frequently used.

LLMs have also been used for MT evaluation: \citet{kocmi23large} showed that GPT-based metrics reach state-of-the-art correlations with human judgments on WMT benchmarks. The benefits of LLM reasoning for this task are still under investigation: \citet{larionov2025deepseek} found that reasoning-enabled LLMs such as DeepSeek-R1 and OpenAI’s o3-mini do not always outperform their non-reasoning counterparts in terms of alignment with human judgments of MT quality.

\section{Evaluation Data and Metrics}
\label{sec:data-metrics}

\begin{table*}[ht]
    \centering
    \begin{tabular}{|p{4.5cm}|p{4.7cm}|p{4.9cm}|} 
    \hline
    & \textbf{Anaphora} & \textbf{Lexical~choice} \\
    \hline
         \textbf{EN context}& It's been a while since I last went to the \textit{river}. & And then the \textit{attack} took place.\\
         \textbf{EN current}& It feels great to finally see \textit{it}. & A truly terrible \textit{attack}.\\
         \hline
         \textbf{FR context (given)} & Ça fait longtemps que je n'ai pas été à la \textit{rivière}.& Et puis l'\textit{attaque} a eu lieu.\\
         \textbf{FR current \#1 (correct)} & C'est chouette de \textit{la} voir enfin.& Une \textit{attaque} vraiment affreuse.\\
         \textbf{FR current \#2 (incorrect)} & C'est chouette de \textit{le} voir enfin.& Un \textit{assaut} vraiment affreux.\\
         \hline
    \end{tabular}
    \caption{Contrastive test items for anaphora and lexical choice from DiscEvalMT \citep{bawden-etal-2018-evaluating}.}
    \label{tab:data-example}
    \end{table*}

\subsection{Benchmark Data}

This study uses the DiscEvalMT benchmark \citep{bawden2018thesis,bawden-etal-2018-evaluating}, designed to evaluate challenges in English-to-French translation that arise from two types of inter-sentential dependencies: pronominal anaphora and lexical cohesion.\footnote{\href{https://github.com/rbawden/discourse-mt-test-sets}{\nolinkurl{https://github.com/rbawden}}}  The test cases are manually constructed.  Each test item consists of two English sentences (`context' and `current'), a French translation of the first one, and two alternative French translations of the second one: one is contextually appropriate, preserving coherence across the two sentences, while the other introduces a discourse-level error.  Together, they form a contrastive pair among which a system should distinguish the correct translation.  

An item for testing the translation of pronominal anaphora is shown in Table~\ref{tab:data-example} (center).  The word `river' is translated as `rivière' (feminine), so FR translation \#1 has correct gender agreement between the object pronoun `la' and the antecedent `rivière', while the second one has wrong agreement.  To control for chance, in another test item, `river' is translated as `fleuve' (masculine), making FR translation \#2 correct.  Two additional test items are made with approximate translations of the antecedent (e.g., `piscine' (fem.) or `cinéma' (masc.)).  In these cases, the translation of the second sentence should still use a pronoun of the same gender as the antecedent, to maintain agreement. 

For lexical choice, each item contains a word in the context sentence, which can be translated in several ways.  The correct translation of the second sentence should use the same word as in the reference translation of the first sentence, to maintain consistency.  In Table~\ref{tab:data-example} (right), the word `attack' is translated as `attaque' in the context sentence, so the current sentence should reuse the same word, as in FR translation \#1.  In an additional test item, `attack' is translated by `assaut' in the context sentence, making FR translation \#2 correct.

The anaphora test set includes 200 items (50 sets of four similar context sentences), each with two candidate translations of the second sentence. Similarly, the lexical choice test set includes 200 items with two candidates each.  We  selected the first half of the data for developing the prompts, and kept the second half for final testing.

\subsection{Metrics}
\label{sec:metrics}

In the first part of our study (Section~\ref{sec:contrastive-task}), the task of the LLM is to distinguish the correct translation of the second sentence in each pair from the wrong one (contrastive task). In the second part (Section~\ref{sec:translation-task}) the task of the LLM is to translate the second sentence into French.

\textbf{The contrastive task} is scored by counting the number of times the correct translation was identified.  The LLMs are prompted to output their response either as `Choice: (1)' or as `Choice: (2)'.  This can be preceded by reasoning in case of CoT prompting. Our evaluation script is tolerant to minor formatting differences in the LLM's output: responses are accepted as correct as long as they clearly indicate the right option (1 or 2); if they cannot be parsed, they are counted as incorrect.  The mean accuracy score is the average over all the answers.

To avoid systematic bias in favor of either candidate \#1 or \#2, we present each test item twice, once with the correct option being first, and once second.  Accordingly, we measure consistency, i.e.\ the sensitivity of the model to the position of the correct translation among the two options.  Ideally, the response should not depend on the position. We define inconsistency as the absolute difference between the accuracy score when the correct option is presented first ($\text{ACC}_{\text{correct=1}}$) and the accuracy score when it is presented second ($\text{ACC}_{\text{correct=2}}$), normalized by the sum of these accuracies.  Inconsistency varies from 0, if accuracy is not influenced at all by the position of the correct answer, to 1, e.g.\ if an LLM always answers `1'. 

    \[
    \text{Inc.} = \frac{|\text{ACC}_{\text{correct=1}} - \text{ACC}_{\text{correct=2}}|}{\text{ACC}_{\text{correct=1}} + \text{ACC}_{\text{correct=2}}}
    \]
    
\textbf{For the translation task}, we provide the LLM with the two EN source sentences, as well as the FR translation of the context sentence, and ask it to translate the second EN sentence (possibly outputting reasoning).  We parse the output to isolate the translation, and score it by comparing it to the correct translation using four metrics: BLEU, chrF, BERTscore and COMET.  The first two capture surface-level overlap, while BERTScore and COMET assess semantic similarity using embeddings.  All metrics are computed using their official Python implementations available via the \texttt{sacrebleu}\footnote{\url{https://github.com/mjpost/sacrebleu}}, \texttt{bert-score}\footnote{\url{https://github.com/Tiiiger/bert_score}}, and \texttt{unbabel-comet}\footnote{\url{https://github.com/Unbabel/COMET}} packages. BLEU and chrF are calculated with the default configurations of \texttt{sacrebleu}. BERTScore is computed using the default multilingual model (\texttt{bert-base-multilingual-cased}) and the language parameter set to French (\texttt{lang='fr'}). COMET uses the \texttt{Unbabel/wmt22-comet-da} model trained on human-annotated translations, 
which was shown to correlate strongly with human judgments.  Given the high density of pronouns, we believe that scoring the entire sentence with frequently-used global metrics is a better solution than comparing pronouns with the APT metric \citep{miculicich-2017-validation}.

\section{Evaluated LLMs and Baseline}
\label{sec:llms}

We evaluated 12 LLMs across different model families and scales. We evaluated GPT models from OpenAI -- GPT-3.5-turbo, GPT-4, GPT-4-turbo, and GPT-4o -- accessed through API requests.\footnote{\url{https://platform.openai.com}} 
We evaluated the following open-weight models: Mistral (7.25B), Phi-4 (14.7B), three versions of Llama (3.1 with 8B, 3.2 with 3.2B, and 3.3 with 70B parameters), and three versions of DeepSeek-R1 (8B, 14B, and 32B); the latter are trained to reason by default, as discussed in Section~\ref{sec:contrastive-task} below.  We do not report results with the Tower LLM \citep{alves2024tower}, although it is a model that was fine-tuned specifically for translation.  However, Tower was not instructed for CoT prompting, and we obtained low scores with it in preliminary experiments.

The LLMs were run locally using the \texttt{Ollama} framework,\footnote{\url{https://ollama.com}} which provides its own versions of the above models.  
For each model, we used the default quantized versions provided by Ollama (Q4\_0 or Q4\_K\_M).
Experiments were done on a Linux server with four NVIDIA RTX 2080~Ti GPUs with 11~GB VRAM each, using parallel querying to maximize throughput, with up to 16 simultaneous query threads.  This limit was determined empirically to ensure stable and efficient throughput.

We include for comparison an encoder-decoder neural MT system, namely the distilled NLLB-200 multilingual model with 600M parameters \citep{nllbteam2022}.  The model was queried using the Transformers library from Hugging Face.\footnote{\url{https://huggingface.co/facebook/nllb-200-distilled-600M}.}  For both tasks (contrastive and translation), we used prefix decoding: the French translation of the first sentence was added as a fixed prefix to the decoder.  Then, for the contrastive task, we generate the translation of the second sentence, and compare it to each of the two options in the test pair using BERTScore. We assume that the most similar one is the model's answer.  For the translation task, the MT model also uses prefix decoding to generate the translation of the second sentence, which is then evaluated.

\section{Contrastive Task}
\label{sec:contrastive-task}

\subsection{Prompts: Reasoning or Not}
\label{sec:contrastive-prompts}

Prompts to LLMs are typically made of a \textit{system prompt}, which indicates the role, persona, or style expected for the answer, and a \textit{user prompt}, which specifies the task, ending with data for the task.  Here, we instruct the LLMs to solve the contrastive task, i.e.\ select the correct translation, with several variants for the system and user prompts.

The \textit{system prompt} can be either \textit{empty} (no system prompt), \textit{simple}, or \textit{detailed}. Each of these versions is shown in Appendix~\ref{sec:prompt-variants-contrastive}.  The detailed version is twice longer than the simple one and contains more constraints, but neither of them gives any instruction on how to solve the task.  The system prompts are the same for the two benchmarks, anaphora or lexical cohesion, on the contrastive task.

The \textit{user prompt} can be either a \textit{simple} definition of the task, which includes the data, or instructions on how to reason \textit{step-by-step} to solve it.  These prompts are also shown in Appendix~\ref{sec:prompt-variants-contrastive}.  The step-by-step reasoning instructions for the anaphora and the lexical choice tasks are very similar: the main difference is at the third step.  For anaphora, this is ``Find the text in English line 1 to which the text found at Step 2 refers'' while for lexical cohesion this is ``Find the text in English line 1 which is identical to the text found at Step 2''.\footnote{These instructions are oriented towards solving the anaphora or lexical cohesion problems, and are therefore more focused than the document-level step-by-step translation considered by \citet{briakou-EtAl:2024:WMT}.}

We designed the prompts through a series of experiments on half of the benchmark data (the development set) with 12 LLMs.  We settled on four combinations of system and user prompts: 
(1)~no system prompt and simple user prompt; (2)~simple system and user prompts; (3)~detailed system prompt and simple user prompt; (4)~simple system prompt and step-by-step (reasoning) user prompt.

The DeepSeek-R1 models are ``reasoning'' models, as they were trained to generate a detailed solution with multiple verifications and to conclude with the final answer.  
The usage instructions for DeepSeek-R1, which we follow, recommend to avoid a system prompt, but to include the words ``please reason step by step'' and indicate the explicit markup of the answer in the user prompt. 

\begin{figure*}[ht]
    \centering
    \includegraphics[width=0.85\textwidth]{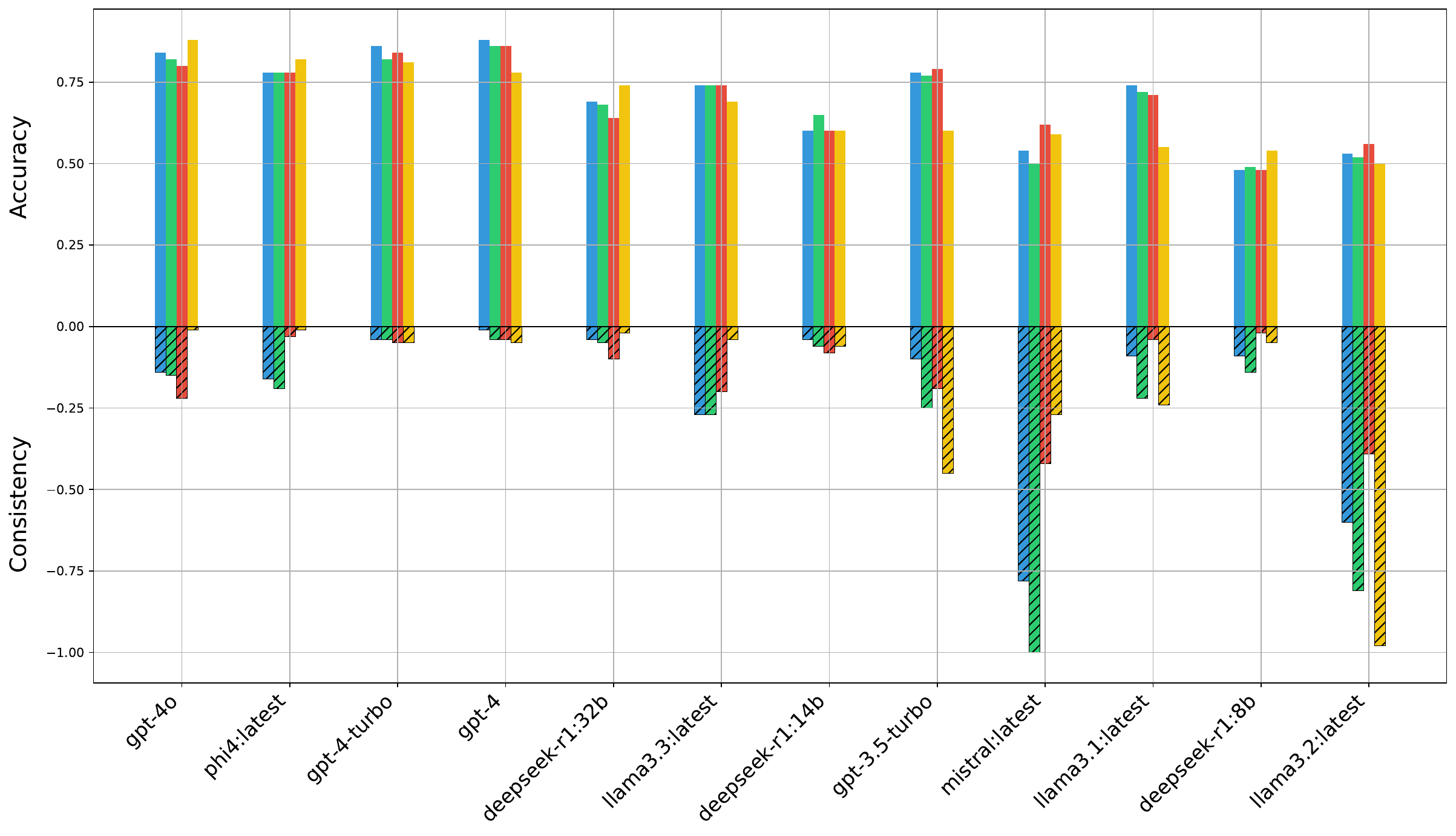}
    \caption{Mean accuracy (top bars) and inconsistency (bottom bars, smaller are better) on the contrastive anaphora task, validation set.  For each LLM, the four bars are those of the prompts: 
    (1)~no system prompt and simple user prompt; (2)~simple system and user prompts; (3)~detailed system prompt and simple user prompt; (4)~simple system prompt and step-by-step (reasoning) user prompt.
    LLMs are ranked by decreasing scores obtained with the last prompt (yellow), which reaches highest overall performance.}
    \label{fig:contrastive-anaphora-results}
\end{figure*}

\begin{figure*}[ht]
  \centering
  \vspace{0.5em}
  \includegraphics[width=.85\textwidth]{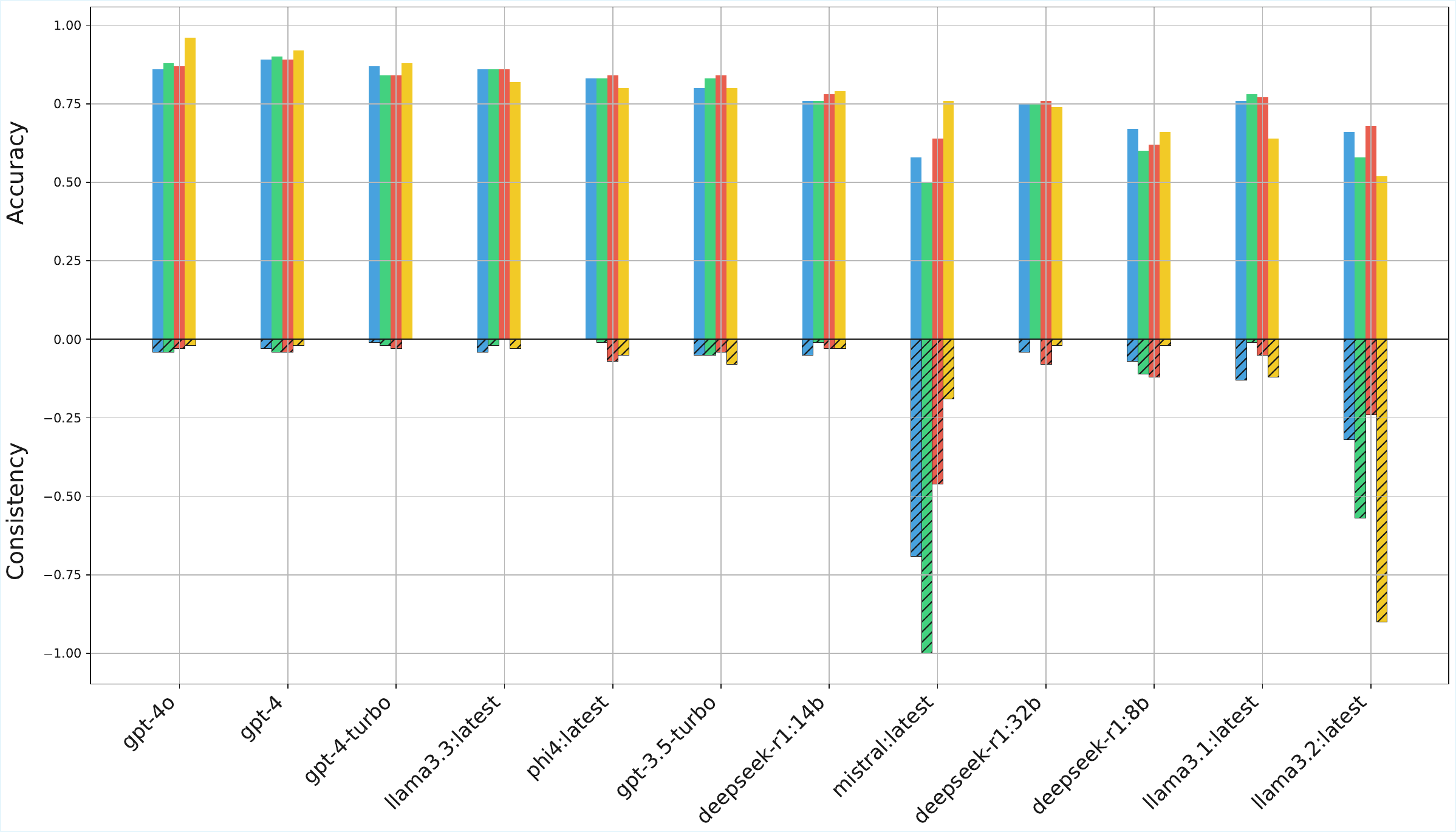}
  \caption{Mean accuracy (top bars) and inconsistency (bottom bars, smaller are better) across prompt configurations for each model on the lexical choice validation set.}
  \label{fig:contrastive-lexical-results}
\end{figure*}

\subsection{Results of the Contrastive Task}
\label{sec:contrastive-results}

The accuracy and inconsistency scores of these combinations of prompts are shown in Figure~\ref{fig:contrastive-anaphora-results} on the anaphora task, for all 12 LLMs.  The numbers are given in Table~\ref{tab:contrastive-anaphora-results} for the five most interesting LLMs, on the development and test sets.  The systems are ranked by decreasing accuracy of the best overall prompt, which is the one with simple system prompt and reasoning user prompt.  The findings from experiments with the validation sets are confirmed on the test sets.  Similarly, the scores for the lexical choice task are shown in Figure~\ref{fig:contrastive-lexical-results}, and the scores for five LLMs on both subsets of the benchmark data are shown in Table~\ref{tab:contrastive-lexical-results}.

\begin{table}[ht]
\centering
\small
\begin{tabular}{|l|c|c|c|c|}
\hline
 & \multicolumn{2}{c|}{\bf Development} & \multicolumn{2}{c|}{\bf Test}\\
\textbf{Model} & \textbf{Acc.↑} & \textbf{Inc.↓} & \textbf{Acc.↑} & \textbf{Inc.↓} \\
\hline
nllb-200 & 0.68 & -- & 0.64 & -- \\
\hline
gpt-4o
 & 0.84 & 0.14 & 0.90 & 0.07 \\
 &  0.82 & 0.15 & 0.92 & 0.06 \\
 &  0.80 & 0.22 & 0.88 & 0.12 \\
 &  \textbf{0.88} & \textbf{0.01} & \textbf{0.97} & 0.01 \\
\hline
gpt-4  & \textbf{0.88} & \textbf{0.01} & 0.94 & 0.04 \\
  & 0.86 & 0.04 & 0.95 & 0.01 \\
  & 0.86 & 0.04 & 0.90 & 0.01 \\
  & 0.78 & 0.05 & 0.88 & 0.00 \\
\hline
gpt-4-  & 0.86 & 0.04 & 0.91 & 0.00 \\
turbo  & 0.82 & 0.04 & 0.88 & 0.07 \\
  & 0.84 & 0.05 & 0.88 & 0.09 \\
  & 0.81 & 0.05 & 0.84 & 0.03 \\
\hline
Phi-4   & 0.78 & 0.16 & 0.90 & 0.01 \\
(14B)   & 0.78 & 0.19 & 0.92 & 0.01 \\
  & 0.79 & 0.04 & 0.89 & 0.04 \\
  & 0.82 & 0.01 & 0.87 & 0.03 \\
\hline 
 DeepSeek-R1  & 0.69 & 0.04 & 0.81 & 0.02 \\
 (32B)  & 0.68 & 0.05 & 0.80 & 0.02 \\
  & 0.64 & 0.10 & 0.68 & 0.02 \\
  & 0.74 & 0.02 & 0.81 & 0.00 \\
\hline
\end{tabular}
\caption{Comparison of accuracy and inconsistency on development and test sets for the \textit{anaphora task}. For each LLM, the four lines of scores are those of the prompts: 
(1)~no system prompt and simple user prompt; (2)~simple system and user prompts; (3)~detailed system prompt and simple user prompt; (4)~simple system prompt and step-by-step (reasoning) user prompt.}
\label{tab:contrastive-anaphora-results}
\end{table}


\begin{table}[ht]
\centering
\small
\begin{tabular}{|l|l|l|c|c|c|c|}
\hline
 & \multicolumn{2}{c|}{\bf Development} & \multicolumn{2}{c|}{\bf Test}\\
\textbf{Model} & \textbf{Acc.↑} & \textbf{Inc.↓} & \textbf{Acc.↑} & \textbf{Inc.↓} \\
\hline
nllb-200 & 0.68 & -- & 0.54 & -- \\ 
\hline
gpt-4o 
 &  0.86 & 0.04 & 0.94 & 0.03 \\
 &  0.88 & 0.04 & 0.94 & 0.04 \\
 &  0.87 & 0.03 & 0.94 & 0.02 \\
 &  \textbf{0.96} & 0.02 & \textbf{0.96} & 0.01 \\
\hline
gpt-4-
 &  0.87 & 0.01 & 0.92 & 0.03 \\
turbo  & 0.84 & 0.02 & 0.92 & 0.03 \\
 &  0.84 & 0.03 & 0.92 & 0.04 \\
 &  0.88 & 0.00 & 0.94 & 0.02 \\
\hline
Phi-4
 &  0.83 & 0.00 & 0.89 & 0.01 \\
(14B)  & 0.83 & 0.01 & 0.91 & 0.02 \\
  & 0.84 & 0.07 & 0.90 & 0.03 \\
  & 0.80 & 0.05 & 0.92 & 0.02 \\
\hline
gpt-4
  & 0.89 & 0.03 & 0.94 & 0.01 \\
  & 0.90 & 0.04 & 0.92 & 0.02 \\
  & 0.89 & 0.04 & 0.92 & 0.03 \\
  & 0.92 & 0.02 & 0.91 & 0.02 \\
\hline
DeepSeek-R1  & 0.76 & 0.05 & 0.86 & 0.04 \\
(14B)   & 0.76 & 0.01 & 0.88 & 0.03 \\
  & 0.78 & 0.03 & 0.84 & 0.07 \\
  & 0.79 & 0.03 & 0.78 & 0.01 \\
\hline
\end{tabular}
\caption{Comparison of accuracy and inconsistency on validation and test sets for the \textit{lexical choice} task. For each LLM, the four lines correspond to the four configurations of prompts.}
\label{tab:contrastive-lexical-results}
\end{table}


To the best of our knowledge, the performance of previous systems is as follows.  The creators of DiscEvalMT designed a system which encodes and decodes jointly the pairs of sentences, reaching accuracies of 0.72 and 0.57 respectively on anaphora and lexical cohesion \citep[Table~2]{bawden-etal-2018-evaluating}.  In a comparative evaluation study, \citet[Table~7]{lopes-etal-2020-document} found that the best performing approach was the concatenation of both sentences on both sides, as proposed by \citet{tiedemann-scherrer-2017-neural}, reaching 0.82 and 0.55 accuracies.  The encoder-decoder model proposed by \citet[Table~4]{pal-etal-2024-document}, which used a separate encoder for the context, reached 0.54 and 0.52 accuracy, while \citet[Table~5]{zhang-etal-2022-multilingual} announced 0.64 accuracy on the anaphora task.  A quality estimation approach with the COMET-QE metric \citep{vernikos-etal-2022-embarrassingly} reached respectively 0.83 and 0.68 accuracy. Finally, our own use of NLLB-200 only reached, respectively, 0.66 and 0.61 accuracy (first line of Table~\ref{tab:contrastive-anaphora-results}).

\textbf{On the anaphora task} (Figure~\ref{fig:contrastive-anaphora-results} and Table~\ref{tab:contrastive-anaphora-results}), the best results on the validation set came from GPT-4o with the reasoning prompt and from GPT-4 with the simplest prompt (no reasoning), both reaching 0.88 accuracy.  However, GPT-4o with reasoning outperformed GPT-4 on the test set, reaching a nearly perfect accuracy of 0.97.  For both systems, the presentation order of the alternatives had almost no influence on the answer, with an inconsistency score of only 0.1.  GPT-4 also did well with the other prompts, except the reasoning one, scoring up to 0.86 with an inconsistency of 0.04, and GPT-4-turbo followed closely. GPT-3.5-turbo showed a larger variability, with inconsistency scores up to 0.45 on the validation set. 

Among LLMs run locally, Phi-4 stood out with 0.82 accuracy with the reasoning prompt and very stable outputs (inconsistency of 0.01), making it the most reliable open-source model. DeepSeek-R1 32B peaked at 0.74 accuracy also with a low inconsistency of 0.02, but smaller DeepSeek variants like the 14B and 8B models dropped to around 0.60 and were less consistent.  Because Deepseek-R1 are reasoning models, the reasoning prompt did not bring significant benefits, apart from a small increase likely related to the instruction to ``reason step-by-step'' recommended by its authors.  As expected from the usage instructions, a detailed system prompt was detrimental.

At the lower end, models like Llama 3.2 and Mistral struggled on the validation set, and were not tested any further. Llama 3.2 was totally unable to understand the reasoning prompt, reaching only 0.50 accuracy (random level) and an inconsistency of 0.98. Mistral scored between 0.54 and 0.59, with inconsistencies over 0.75 in several cases. 

Overall, higher accuracy goes hand-in-hand with better consistency. The reasoning prompt was helpful for models that could handle it. For smaller or less instructed models, more complex prompts often led to worse outcomes or formatting issues.

\textbf{On the lexical choice task} (Figure~\ref{fig:contrastive-lexical-results} and Table~\ref{tab:contrastive-lexical-results}) the trends are similar.  The scores are even higher than for anaphora, with excellent consistency.  The best performance came from GPT-4o with the reasoning-style prompt, reaching an accuracy of 0.96 and a very low inconsistency of 0.02. GPT-4 also performed strongly, with up to 0.92 accuracy, although its results varied slightly more across prompts.  Again, models with strong reasoning capabilities, such as GPT-4o and GPT-4, consistently outperformed others, especially with the CoT prompt, and among open-source options, Phi-4 offered the best combination of performance and efficiency.

In terms of the length of response, which correlates with cost, DeepSeek-R1 models consistently generated long responses (over 600 tokens), regardless of the prompt, as they are trained to perform CoT reasoning.  In contrast, most other models had output lengths that were more clearly shaped by the prompt. Llama 3.3 and Phi-4 produced some of the longest reasoning responses apart from DeepSeek-R1, with 279 and 227 tokens on average under the reasoning prompt. On the contrary, GPT-3.5-turbo, Mistral, and Llama 3.2 generated very short outputs, even with reasoning prompts, rarely exceeding 10 tokens.  Therefore, given the higher cost of CoT responses, reasoning prompts are only worth using when they improve performance, as in the case of GPT-4o and Phi-4. 

\section{Translation Task}
\label{sec:translation-task}

We now test the ability of LLMs to correctly translate a sentence when inter-sentential constraints -- pronominal anaphora and lexical cohesion -- are involved.  Indeed, as observed by \citet{post-junczysdowmunt:2024:WMT} for encoder-decoder MT systems, scores on the contrastive task are not necessarily correlated with translation quality.

Using the same dataset, we give each LLM the two source sentences (EN) and the reference translation of the first sentence (FR). We obtain the translation of the second sentence from the LLM and score it against the reference translation present in the dataset using the four evaluation metrics presented in Section~\ref{sec:metrics} above: BLEU, chrF, BERTScore, and COMET.  As prompt engineering relied less on the development set than for the first task, we report below the results on the full set of 200 examples for each task.


\subsection{Prompts: Reasoning or Not}
\label{sec:translation-prompts}

Our goals are again to determine the best performance of LLMs on the two benchmarks, and to find if CoT reasoning improves performance over translation with no reasoning.  The no-reasoning prompt is the same for both benchmarks, as shown in Appendix~\ref{sec:prompt-variants-translation}. The system part contains general instructions for translation, while the user part simply provides the two EN sentences and the FR translation of the first one.

For the reasoning prompts (also in Appendix~\ref{sec:prompt-variants-translation}), we make explicit in the system prompt several steps that guide the reasoning for each task. An additional instruction at the end requires the LLM to enclose the reasoning between XML-like tags, which leads to an easier to parse output, as well as similar or better performance.

\begin{table*}[ht]
\centering
\small
\begin{tabular}{|l|ccc|ccc|ccc|ccc|}
\hline
 &
\multicolumn{3}{c|}{\textbf{BLEU}} &
\multicolumn{3}{c|}{\textbf{chrF}} &
\multicolumn{3}{c|}{\textbf{BERTScore}} &
\multicolumn{3}{c|}{\textbf{COMET}} \\
\textbf{Model} & w/o & w/ & $\Delta$ & w/o & w/ & $\Delta$ & w/o & w/ & $\Delta$ & w/o & w/ & $\Delta$ \\
\hline
nllb-200         & 37 & -- & -- & 60 & -- & -- & .90 & -- & -- & 0.87 & -- & -- \\
gpt-4            & \textbf{49} & 53 & +3.78 & \textbf{70} & \textbf{71} & +1.77 & \textbf{.92} & \textbf{.93} & +0.0033 & \textbf{.92} & \textbf{.92} & +0.0006 \\
gpt-4o           & \textbf{49} & \textbf{54} & +5.35 & \textbf{70} & \textbf{71} & +1.96 & \textbf{.92} & .92 & +0.0016 & .91 & .91 & +0.0017 \\
gpt-4-turbo      & 45 & 49 & +4.17 & 67 & 67 & +0.90 & .92 & .90 & -0.0173 & .91 & .90 & -0.0084 \\
gpt-3.5-turbo    & 44 & 49 & +5.22 & 66 & 68 & +2.15 & .91 & .91 & +0.0017 & .91 & .90 & -0.0080 \\
Llama 3.3        & 44 & 47 & +2.68 & 66 & 67 & +1.08 & .92 & .91 & -0.0046 & .90 & .90 & -0.0043 \\
Phi-4            & 43 & 49 & \textbf{+5.58} & 64 & 68 & \textbf{+4.39} & .91 & .92 & \textbf{+0.0096} & .88 & .91 & \textbf{+0.0270} \\
DeepSeek-R1 32B  & 35 & 39 & +4.34 & 59 & 59 & -0.04 & .89 & .84 & -0.0551 & .87 & .84 & -0.0286 \\
Llama 3.1        & 34 & 30 & -3.78 & 58 & 54 & -3.35 & .89 & .87 & -0.0213 & .86 & .85 & -0.0130 \\
DeepSeek-R1 14B  & 34 & 33 & -1.06 & 58 & 53 & -4.91 & .89 & .79 & -0.0927 & .86 & .79 & -0.0647 \\
Mistral          & 27 & 27 & -0.03 & 51 & 50 & -0.80 & .86 & .85 & -0.0084 & .82 & .80 & -0.0182 \\
Llama 3.2        & 27 & 23 & -3.48 & 51 & 47 & -3.83 & .87 & .83 & -0.0390 & .82 & .76 & -0.0579 \\
DeepSeek-R1 8B   & 24 & 21 & -2.48 & 50 & 45 & -4.75 & .86 & .77 & -0.0908 & .80 & .77 & -0.0298 \\
\hline
\end{tabular}
\caption{Translation quality scores on the \textit{anaphora benchmark}: without reasoning (w/o), with reasoning (w/), and difference ($\Delta$) between the latter and the former. Values for w/o and w/ are rounded to 2 digits, and $\Delta$ values to 4. Positive values of $\Delta$ indicate progress due to reasoning.}
\label{tab:translation-anaphora-results}
\end{table*}

\begin{table*}[ht]
\centering
\small
\begin{tabular}{|l|ccc|ccc|ccc|ccc|}
\hline
 &
\multicolumn{3}{c|}{\textbf{BLEU}} &
\multicolumn{3}{c|}{\textbf{chrF}} &
\multicolumn{3}{c|}{\textbf{BERTScore}} &
\multicolumn{3}{c|}{\textbf{COMET}} \\
\textbf{Model} & w/o & w/ & $\Delta$ & w/o & w/ & $\Delta$ & w/o & w/ & $\Delta$ & w/o & w/ & $\Delta$ \\
\hline
nllb-200         & 31 & -- & -- & 51 & -- & -- & .86 & -- & -- & 0.77 & -- & -- \\
gpt-4o           & 54 & 54 & +0.31 & 70 & 69 & -0.26 & .92 & .92 & -0.0015 & .89 & .88 & -0.0054 \\
gpt-4            & 51 & 54 & +2.58 & 67 & 69 & +2.16 & .91 & .92 & +0.0051 & .86 & .87 & +0.0070 \\
gpt-4-turbo      & 50 & 49 & -0.31 & 66 & 65 & -0.83 & .91 & .89 & -0.0209 & .87 & .86 & -0.0126 \\
gpt-3.5-turbo    & 47 & 49 & +2.11 & 65 & 66 & +0.81 & .91 & .91 & +0.0041 & .86 & .86 & -0.0059 \\
Llama3.3         & 46 & 47 & +0.15 & 62 & 63 & +0.90 & .90 & .90 & -0.0018 & .85 & .84 & -0.0030 \\
Phi-4            & 43 & 44 & +1.77 & 60 & 61 & +1.74 & .89 & .90 & +0.0064 & .83 & .84 & +0.0021 \\
DeepSeek-R1 32B  & 39 & 41 & +2.07 & 56 & 58 & +2.01 & .88 & .88 & -0.0011 & .82 & .81 & -0.0054 \\
DeepSeek-R1 14B  & 38 & 33 & -5.10 & 55 & 51 & -3.99 & .87 & .85 & -0.0226 & .80 & .77 & -0.0330 \\
Llama3.1         & 35 & 29 & -5.71 & 52 & 49 & -3.53 & .87 & .86 & -0.0144 & .79 & .77 & -0.0261 \\
DeepSeek-R1 8B   & 31 & 30 & -0.63 & 48 & 48 & -0.28 & .86 & .83 & -0.0227 & .76 & .75 & -0.0115 \\
Mistral          & 27 & 28 & +1.17 & 45 & 45 & +0.11 & .84 & .85 & +0.0046 & .74 & .75 & +0.0101 \\
Llama3.2         & 25 & 22 & -2.81 & 42 & 43 & +0.63 & .83 & .82 & -0.0121 & .73 & .70 & -0.0256 \\
\hline
\end{tabular}
\caption{Translation quality scores on the \textit{lexical choice benchmark}: without reasoning (w/o), with reasoning (w/), and difference ($\Delta$) between the latter and the former. Values for w/o and w/ are rounded to 2 digits, and $\Delta$ values to 4. Positive values of $\Delta$ indicate progress due to reasoning.}
\label{tab:translation-lexical-results}
\end{table*}

\subsection{Results of the Translation Task}
\label{sec:translation-results}

\textbf{On the anaphora benchmark}, the translation scores with or without reasoning, and their differences ($\Delta$), are shown in Table~\ref{tab:translation-anaphora-results}.  The results show that the GPT family delivers the strongest performance, and the largest increase when reasoning is encouraged. The smaller ``turbo'' versions benefit most from reasoning in terms of BERTScore and COMET.  Phi-4 exhibits the highest improvement on all metrics when reasoning is used.  Together with Llama 3.3, they are the highest scoring open-weight models, though Llama 3.3 draws only modest benefits from reasoning.  The DeepSeek-R1 models present a mixed picture: only the largest model (32B) improves its BLEU score, though not the other scores.  

Models with weaker baselines, i.e., DeepSeek-R1~8B and~14B, Llama~3.1 and~3.2, and Mistral decrease across all metrics when prompted to reason.  The scores of NLLB-200 place it midway, behind GPT, Llama~3.3 and Phi, but ahead of DeepSeek-R1 and Mistral.   A possible explanation for the scores of DeepSeek-R1 is that, being a reasoning model, the additional reasoning instructions in the user prompt (beyond ``please reason step by step'') have a detrimental effect, together with the system prompt, especially for the smaller models.

\textbf{On the lexical choice benchmark}, the ranking of the LLMs is similar to the one on anaphora (see Table~\ref{tab:translation-lexical-results}). Within the GPT family, reasoning improves GPT-4, GPT-4o, and GPT-3.5-turbo across most metrics.  While GPT-4-turbo shows an unexpected small decline, GPT-4 seems to make a slightly more effective use of the reasoning instructions, despite GPT-4o's stronger baseline score.  Among open-weight models, Phi-4 again improves on all metrics with the CoT prompt, while LLama~3.3 and DeepSeek-R1~32B improve only in BLEU and chrF, Mistral gains slightly, and the other models degrade, while NLLB-200 is close to the bottom of the ranking.  Lexical choice may pose a more difficult challenge compared to pronominal anaphora, as acceptable translations of ambiguous words are more numerous.  This may explain why improvements through reasoning are slightly less consistent for the lexical choice task.

\subsection{The ``Wise Get Wiser'' Effect}
\label{sec:wise-get-wiser}

In both the anaphora and lexical choice experiments, we observe a surprising effect.  Unlike many techniques which tend to improve weaker models but do not benefit the top-scoring ones, here reasoning is more beneficial to the LLMs which already score highest without it.  Table~\ref{tab:ana-gen-wise-bar} shows the Pearson correlations and the Spearman rank correlations between baseline scores without reasoning and the improvements obtained with reasoning ($\Delta$), for both tasks.  For anaphora, high coefficients for all metrics (0.59--0.81) confirm a consistent ``wise get wiser'' pattern: models that start from higher baselines tend to benefit more from reasoning.  The effect is weaker for lexical choice, although correlations remain positive (0.21--0.52).

\begin{table}[b]
\centering
\setlength{\tabcolsep}{4pt}
\begin{tabular}{|l|l|cccc|}
\hline
 & & \textbf{BLEU} & \textbf{chrF} & \textbf{BERT} & \textbf{COMET} \\
  & &  &  & \textbf{Score} &  \\
\hline
\textbf{Anaphora} & \textit{P} & 0.81 & 0.76 & 0.60 & 0.59 \\
         & \textit{S} & 0.68 & 0.72 & 0.64 & 0.74 \\
\hline
\textbf{Lexical}  & \textit{P} & 0.40 & 0.21 & 0.31 & 0.30 \\
\textbf{choice}   & \textit{S} & 0.52 & 0.27 & 0.30 & 0.25 \\
\hline
\end{tabular}
\caption{Pearson (\textit{P}) and Spearman (\textit{s}) correlations between the baseline scores without reasoning and the improvements due to reasoning ($\Delta$).  The large correlations show the ``wise get wiser'' effect.}
\label{tab:ana-gen-wise-bar}
\end{table}

\begin{figure}[t]
  \centering
  \includegraphics[width=.97\linewidth]{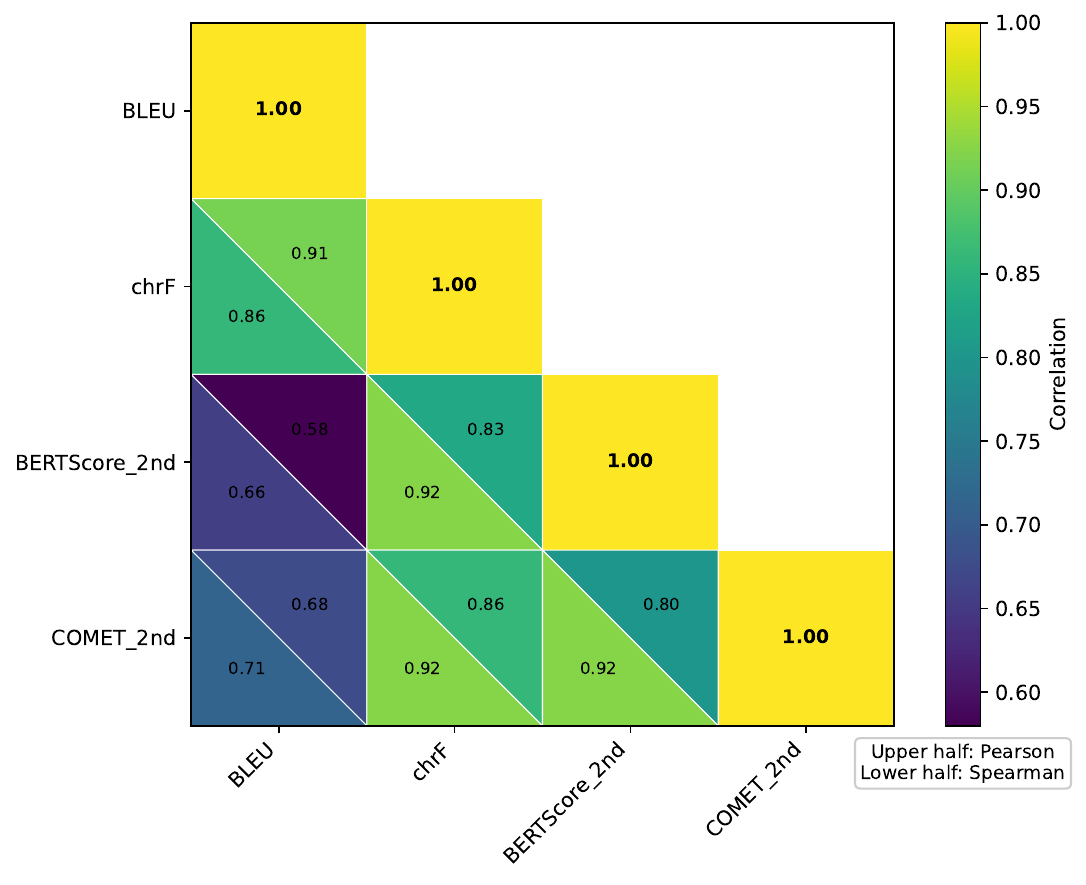}
    \includegraphics[width=.97\linewidth]{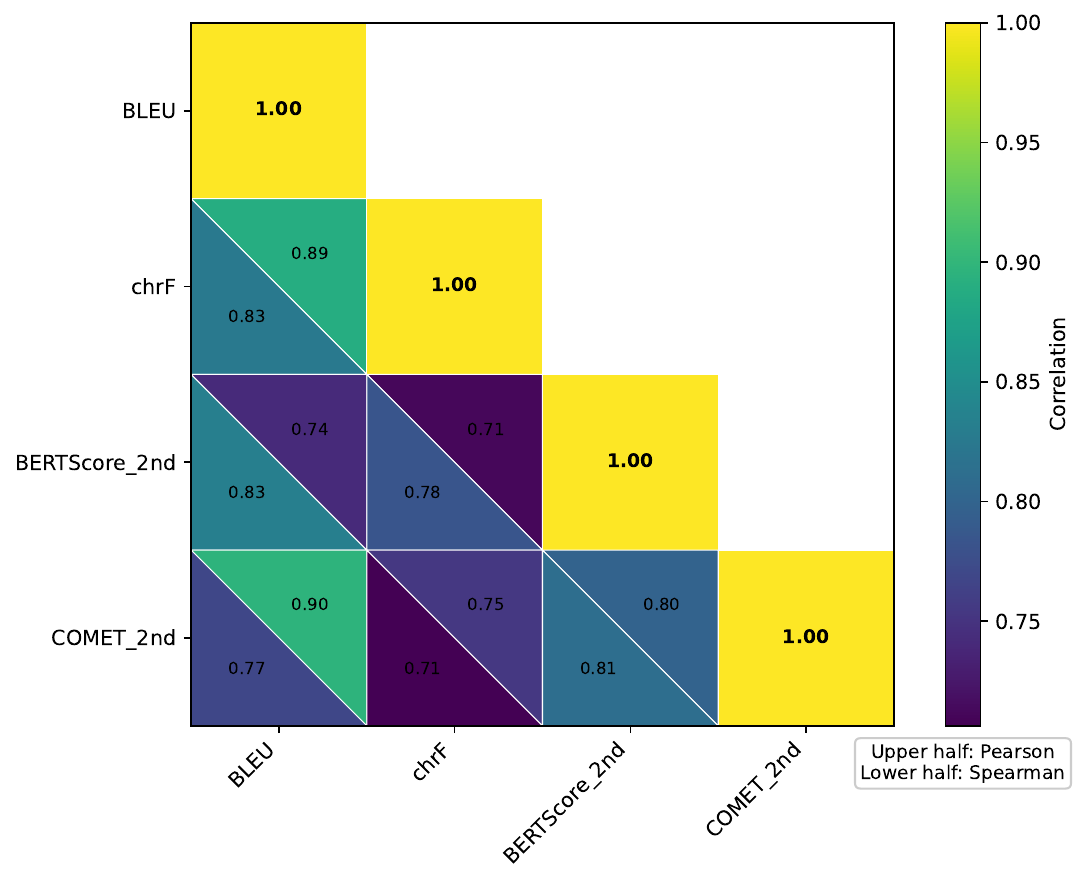}
  \caption{Correlations between the $\Delta$ values (reasoning minus no-reasoning) obtained with several MT metrics for 12 LLMs, on anaphora translation (upper part) and lexical choice translation (lower part). Each cell shows Pearson (upper triangle) and Spearman (lower triangle) correlations.}
  \label{fig:ana-gen-corr-new} 
\end{figure}

Moreover, we confirm that the variations of translation scores between the prompts without reasoning and those with reasoning are consistent across all metrics.  This is important because scoring LLM translations uses MT metrics which correlate imperfectly with human preferences, unlike the accuracy metric of the first task.  Figure~\ref{fig:ana-gen-corr-new}, upper part, shows correlations among the $\Delta$ values of all systems, between all pairs of metrics. The rather large coefficients (Pearson: 0.58--0.91, Spearman: 0.66--0.92) indicate that models scoring well with respect to the others on one metric tend to score well on the other metrics as well.  Similar values are observed for the lexical choice task, shown in Figure~\ref{fig:ana-gen-corr-new}, lower part (Pearson: 0.71--0.90, Spearman: 0.71.--0.83).

\section{Conclusion}
\label{sec:conclusion}

This paper evaluated the capacity of several LLMs to pass two benchmarks for contextual MT, one targeting pronoun translation and the other one targeting lexical coherence.  We compared several types of prompts, and showed that CoT prompting elicits reasoning steps that lead to the best results.  We found that the best models improve considerably the state of the art on the contrastive task, with accuracies slightly above 0.95 for both targeted phenomena.  They also score highly on the four translation quality metrics, with COMET scores of 0.92 and 0.89.  Moreover, we observed a ``wise get wiser'' effect, as the improvement brought by reasoning is positively correlated to the scores of the same LLMs without reasoning.  In other words, the strongest models are also those that benefit the most from reasoning.

These results point to a possible future solution for improving translation with LLMs thanks to reasoning.  The solution would need first to identify locations in documents where reasoning is likely to be beneficial, then generate the reasoning that makes translation choices explicit, separating it with markup from the actual translation.  Either a generic CoT prompt could be used, or a specific one could be applied depending on the identified difficulties -- as we did here with slightly different prompts for anaphora and lexical choice.  Such a self-reflecting behavior could lend itself naturally to an agentic AI approach, in which a first-pass translation generated without reasoning could be improved by explicitly solving inter-sentential dependencies.


\section{Acknowledgments}
We are grateful to the Swiss National Science Foundation for its support through the EXOMAT grant n.\ 228494, External Knowledge for Low-resource Machine Translation.
We acknowledge the support of HES-SO through the grant n.\ AGP-140146.
We would like to warmly thank the anonymous LREC reviewers for their insightful suggestions.


\section{Bibliographical References}
\label{sec:reference}\vspace{-2em}
\bibliographystyle{lrec2026-natbib}
\bibliography{references.bib}


\onecolumn 
\appendix
\section*{Appendix}


\section{Contrastive Task}
\label{sec:contrastive-task-appendix}

\subsection{Prompt Variants}
\label{sec:prompt-variants-contrastive}

This section presents the full text of the prompts used in our \textit{contrastive} evaluation.  
Following common prompt engineering practice, we have experimented with various system prompts (or none) and user prompts. The various prompts differ in their structures and instruction levels, while maintaining a consistent task format.  
Placeholders appear in curly braces in the prompt templates: \texttt{\{context\_en\}}, \texttt{\{source\_sentence\}}, \texttt{\{context\_fr\}}, \texttt{\{option1\}}, \texttt{\{option2\}}.
In the contrastive setting, two candidate translations are presented, and the model must output a constrained choice, either as ``Choice: 1'' or as ``Choice: 2''. 

\paragraph{No system prompt, simple user prompt} \mbox{}\\[0.5em]
\indent\textit{User prompt}\\
{\ttfamily\footnotesize
Here is a short text with two sentences in English:\\
\texttt{\{context\_en\}}\\
\texttt{\{source\_sentence\}}\\
Here are two different translations into French:\\
1.\ \texttt{\{context\_fr\}}\\ \texttt{\{option1\}}\\
2.\ \texttt{\{context\_fr\}}\\ \texttt{\{option2\}}\\
Which one is more correct?\\
Please answer \texttt{1} if the first one is more correct or answer \texttt{2} if the second one is more correct. Do not add explanations.\\
MAKE SURE you only answer in the following manner: \texttt{Choice: (1 or 2)}
}

\vspace{0.5em}

\paragraph{Simple system and user prompts}\mbox{}\\[0.5em]
\indent\textit{System prompt}\\
{\ttfamily\footnotesize
You are a language evaluation assistant. Your task is to compare two French translations of an English text and decide which is more correct.\\
When providing your answer, strictly follow this format: \texttt{Choice: (1 or 2)}\\
Do NOT include any explanation or additional text. Only output the specified format.
}

\textit{User prompt}\\
{\ttfamily\footnotesize
Here is a short text with two sentences in English:\\
\texttt{\{context\_en\}}\\
\texttt{\{source\_sentence\}}\\
Here are two different translations into French:\\
1.\ \texttt{\{context\_fr\}}\\ \texttt{\{option1\}}\\
2.\ \texttt{\{context\_fr\}}\\ \texttt{\{option2\}}\\
Which one is more correct?\\
Please answer \texttt{1} if the first one is more correct or answer \texttt{2} if the second one is more correct.
}

\vspace{0.5em}

\paragraph{Detailed system prompt, simple user prompt}\mbox{}\\[0.5em]
\indent\textit{System prompt}\\
{\ttfamily\footnotesize
You are a strict translation evaluation assistant. Your ONLY task is to determine which of two French translations of an English text is more correct.\\
\#\#\# Instructions:\\
-- You will receive an English text containing two sentences.\\
-- You will be provided with two different translations in French.\\
-- Your task is to determine which translation is more accurate.\\
-- You must respond in EXACTLY this format: Choice: (1 or 2)\\
-- You MUST NOT provide any explanations, thoughts, or additional text.\\
-- Any deviation from the required format is strictly prohibited.\\
-- If you understand the instructions, respond ONLY in the required format.
}

\textit{User prompt}\\
{\ttfamily\footnotesize
Here is a short text with two sentences in English:\\
\texttt{\{context\_en\}}\\
\texttt{\{source\_sentence\}}\\
Here are two different translations into French:\\
1.\ \texttt{\{context\_fr\}}\\ \texttt{\{option1\}}\\
2.\ \texttt{\{context\_fr\}}\\ \texttt{\{option2\}}\\
Which one is more correct?
}

\vspace{0.5em}

\paragraph{Simple system prompt, step-by-step (reasoning) user prompt for anaphora}\mbox{}\\[0.5em]
\indent\textit{System prompt}\\
{\ttfamily\footnotesize
You are a helpful assistant. You have to find which translation from English to French is correct. Always keep your answers short.
}

\textit{User prompt (reasoning)}\\
{\ttfamily\footnotesize
Which one is more correct? Please answer \texttt{1} if the first one is more correct or answer \texttt{2} if the second one is more correct.\\
Short text with two lines in English.\\
English line 1: \texttt{\{context\_en\}}\\
English line 2: \texttt{\{source\_sentence\}}\\
French translation of English line 1: \texttt{\{context\_fr\}}\\
French translation number 1 of line 2: \texttt{\{option1\}}\\
French translation number 2 of line 2: \texttt{\{option2\}}\\
Let's reason step by step to find the correct French translation of line 2.\\
1.\ Find the difference between the two translations of line 2.\\
2.\ Find the text in English line 2 which is the cause of the translation difference at Step 1.\\
3.\ Find the text in English line 1 to which the text found at Step 2 refers.\\
4.\ Find how the text from Step 3 is translated in French translation of English line 1.\\
5.\ Find the correct word in French to refer to this text.\\
6.\ Find which French translation includes this word.\\
Select \texttt{1} if French translation number 1 is more correct, or \texttt{2} if French translation number 2 is more correct.\\
When providing your answer, strictly follow this format: \texttt{Choice: (1 or 2)}
}

\vspace{0.5em}

\begin{samepage}
\paragraph{Simple system prompt, step-by-step (reasoning) user prompt for lexical choice}\mbox{}\\[0.5em]
\indent\textit{System prompt}\\
{\ttfamily\footnotesize
You are a helpful assistant. You have to find which translation from English to French is correct. Always keep your answers short.
}

\textit{User prompt (reasoning)}\\
{\ttfamily\footnotesize
Which one is more correct? Please answer \texttt{1} if the first one is more correct or answer \texttt{2} if the second one is more correct.\\
Short text with two lines in English.\\
English line 1: \texttt{\{context\_en\}}\\
English line 2: \texttt{\{source\_sentence\}}\\
French translation of English line 1: \texttt{\{context\_fr\}}\\
French translation number 1 of line 2: \texttt{\{option1\}}\\
French translation number 2 of line 2: \texttt{\{option2\}}\\
Let's reason step by step to find the correct French translation of line 2.\\
1.\ Find the difference between the two translations of line 2.\\
2.\ Find the text in English line 2 which is the cause of the difference at Step 1.\\
3.\ Find the text in English line 1 which is identical to the text found at Step 2.\\
4.\ Find how the text from Step 3 is translated in French translation of English line 1.\\
5.\ Find the correct word in French to refer to this text.\\
6.\ Find which French translation includes this word.\\
Select \texttt{1} if French translation number 1 is more correct, or \texttt{2} if French translation number 2 is more correct.\\
When providing your answer, strictly follow this format: \texttt{Choice: (1 or 2)}
}
\end{samepage}

\FloatBarrier
\subsection{Results: Anaphora}

Below are the results on the \textbf{contrastive anaphora benchmark} for each LLM and prompt type, in terms of accuracy and inconsistency scores, elapsed time and cost of LLM.

\FloatBarrier

{
\setlength{\abovecaptionskip}{8pt}
\begin{table}[ht]
\centering
\small
\setlength{\tabcolsep}{5pt}\renewcommand{\arraystretch}{1.05}
\resizebox{0.95\textwidth}{!}{%
\begin{tabular}{|l|c|c|c|c|}
\hline
 & \multicolumn{4}{c|}{\textbf{System and user prompts}} \\
\textbf{Model} & \textbf{None \& simple} & \textbf{Both simple} & \textbf{Detailed \& simple)} & \textbf{Simple \& step-by-step} \\
\hline
gpt-4o             & 0.84 / 0.14 & 0.82 / 0.15 & 0.80 / 0.22 & 0.88 / 0.01 \\
\hline
Phi-4 (14B)        & 0.78 / 0.16 & 0.78 / 0.19 & 0.79 / 0.04 & 0.82 / 0.01 \\
\hline
gpt-4-turbo        & 0.86 / 0.04 & 0.82 / 0.04 & 0.84 / 0.05 & 0.81 / 0.05 \\
\hline
gpt-4              & 0.88 / 0.01 & 0.86 / 0.04 & 0.86 / 0.04 & 0.78 / 0.05 \\
\hline
DeepSeek-R1 (32B)  & 0.69 / 0.04 & 0.68 / 0.05 & 0.64 / 0.10 & 0.74 / 0.02 \\
\hline
LLaMA 3.3 (70B)    & 0.74 / 0.27 & 0.74 / 0.27 & 0.74 / 0.20 & 0.69 / 0.04 \\
\hline
DeepSeek-R1 (14B)  & 0.60 / 0.04 & 0.65 / 0.06 & 0.60 / 0.08 & 0.60 / 0.06 \\
\hline
gpt-3.5-turbo      & 0.78 / 0.10 & 0.77 / 0.25 & 0.79 / 0.19 & 0.60 / 0.45 \\
\hline
Mistral (7B)       & 0.54 / 0.78 & 0.50 / 1.00 & 0.62 / 0.42 & 0.59 / 0.27 \\
\hline
LLaMA 3.1 (8B)     & 0.74 / 0.09 & 0.72 / 0.22 & 0.71 / 0.04 & 0.55 / 0.24 \\
\hline
DeepSeek-R1 (8B)   & 0.48 / 0.09 & 0.49 / 0.14 & 0.48 / 0.02 & 0.54 / 0.05 \\
\hline
LLaMA 3.2 (3B)     & 0.53 / 0.60 & 0.52 / 0.81 & 0.56 / 0.39 & 0.50 / 0.98 \\
\hline
\end{tabular}}
\caption{Accuracy (\(\uparrow\)) and inconsistency (\(\downarrow\)) across models and prompts on the \textbf{anaphora} validation set (separated by `/' in each cell). The systems are ranked by decreasing accuracy of the best overall prompt, which is the fourth one.}
\label{tab:ana-accuracy-inconsistency-grid}
\end{table}
}

{
\setlength{\abovecaptionskip}{8pt}
\begin{table}[ht]
\centering
\small
\setlength{\tabcolsep}{6pt}\renewcommand{\arraystretch}{1.05}
\resizebox{0.95\textwidth}{!}{%
\begin{tabular}{|l|c|c|c|c|}
\hline
 & \multicolumn{4}{c|}{\textbf{System and user prompts}} \\
\textbf{Model} & \textbf{None \& simple} & \textbf{Both simple} & \textbf{Detailed \& simple)} & \textbf{Simple \& step-by-step} \\
\hline
Phi-4 (14B)        & 0.36  & 0.39  & 0.36  & 2.13  \\
\hline
DeepSeek-R1 (32B)  & 10.02 & 10.26 & 9.15  & 8.49  \\
\hline
LLaMA 3.3 (70B)    & 3.03  & 2.64  & 2.46  & 62.43 \\
\hline
DeepSeek-R1 (14B)  & 6.93  & 5.01  & 4.50  & 6.84  \\
\hline
Mistral (7B)      & 0.30  & 0.27  & 0.27  & 0.39  \\
\hline
LLaMA 3.1 (8B)     & 0.30  & 0.33  & 0.30  & 1.35  \\
\hline
DeepSeek-R1 (8B)   & 4.11  & 4.41  & 4.62  & 3.42  \\
\hline
LLaMA 3.2 (3B)     & 0.27  & 0.27  & 0.33  & 0.42  \\
\hline\hline
gpt-4o             & 0.228 & 0.258 & 0.234 & 0.768 \\
\hline
gpt-4-turbo        & 0.294 & 0.276 & 0.252 & 0.894 \\
\hline
gpt-4              & 0.798 & 0.960 & 1.308 & 3.060 \\
\hline
gpt-3.5-turbo      & 0.192 & 0.240 & 0.198 & 0.204 \\
\hline
\end{tabular}}
\caption{Mean elapsed time per prompt in seconds, across prompt configurations on the \textbf{anaphora} validation set. Open-source (Ollama) models listed first, followed by OpenAI models.}
\label{tab:latency-all-anaphora}
\end{table}
}

{
\setlength{\abovecaptionskip}{8pt}
\begin{table}[ht]
\centering
\small
\setlength{\tabcolsep}{6pt}\renewcommand{\arraystretch}{1.05}
\resizebox{0.95\textwidth}{!}{
\begin{tabular}{|l|c|c|c|c|}
\hline
 & \multicolumn{4}{c|}{\textbf{System and user prompts}} \\
\textbf{Model} & \textbf{None \& simple} & \textbf{Both simple} & \textbf{Detailed \& simple)} & \textbf{Simple \& step-by-step} \\
\hline
gpt-4o         & 0.090086 & 0.109572 & 0.137068 & 0.499464 \\
\hline
gpt-4-turbo    & 0.364640 & 0.431000 & 0.550140 & 1.417670 \\
\hline
gpt-4          & 1.036560 & 1.268280 & 1.604280 & 4.095720 \\
\hline
gpt-3.5-turbo  & 0.017634 & 0.022219 & 0.027188 & 0.037896 \\
\hline
\end{tabular}}
\caption{Total API cost (USD) by prompt configuration for OpenAI models on the \textbf{anaphora} validation set. Experiments were run in summer~2025; reported costs reflect that period’s pricing and our token usage.}
\label{tab:cost-openai-anaphora}
\end{table}
}

{
\setlength{\abovecaptionskip}{8pt}
\begin{table}[ht]
\centering
\small
\setlength{\tabcolsep}{5pt}\renewcommand{\arraystretch}{1.05}
\resizebox{0.95\textwidth}{!}{%
\begin{tabular}{|l|l|l|c|c|c|c|}
\hline
\textbf{Model} & \textbf{System} & \textbf{User} & \textbf{Val. Acc.} & \textbf{Val. Inc.} & \textbf{Test Acc.} & \textbf{Test Inc.} \\
\hline
gpt-4o            & none    & simple        & 0.84 & 0.14 & 0.90 & 0.07 \\
gpt-4o            & simple  & simple        & 0.82 & 0.15 & 0.92 & 0.06 \\
gpt-4o            & detailed& simple        & 0.80 & 0.22 & 0.88 & 0.12 \\
gpt-4o            & simple  & step-by-step  & 0.88 & 0.01 & 0.97 & 0.01 \\
\hline
Phi-4 (14B)       & none    & simple        & 0.78 & 0.16 & 0.90 & 0.01 \\
Phi-4 (14B)       & simple  & simple        & 0.78 & 0.19 & 0.92 & 0.01 \\
Phi-4 (14B)       & detailed& simple        & 0.79 & 0.04 & 0.89 & 0.04 \\
Phi-4 (14B)       & simple  & step-by-step  & 0.82 & 0.01 & 0.87 & 0.03 \\
\hline
gpt-4-turbo       & none    & simple        & 0.86 & 0.04 & 0.91 & 0.00 \\
gpt-4-turbo       & simple  & simple        & 0.82 & 0.04 & 0.88 & 0.07 \\
gpt-4-turbo       & detailed& simple        & 0.84 & 0.05 & 0.88 & 0.09 \\
gpt-4-turbo       & simple  & step-by-step  & 0.81 & 0.05 & 0.84 & 0.03 \\
\hline
gpt-4             & none    & simple        & 0.88 & 0.01 & 0.94 & 0.04 \\
gpt-4             & simple  & simple        & 0.86 & 0.04 & 0.95 & 0.01 \\
gpt-4             & detailed& simple        & 0.86 & 0.04 & 0.90 & 0.01 \\
gpt-4             & simple  & step-by-step  & 0.78 & 0.05 & 0.88 & 0.00 \\
\hline
DeepSeek-R1 (32B) & none    & simple        & 0.69 & 0.04 & 0.81 & 0.02 \\
DeepSeek-R1 (32B) & simple  & simple        & 0.68 & 0.05 & 0.80 & 0.02 \\
DeepSeek-R1 (32B) & detailed& simple        & 0.64 & 0.10 & 0.68 & 0.02 \\
DeepSeek-R1 (32B) & simple  & step-by-step  & 0.74 & 0.02 & 0.81 & 0.00 \\
\hline
\end{tabular}}
\caption{Comparison of accuracy (Acc.\ \(\uparrow\)) and inconsistency (Inc.\ \(\downarrow\)) on validation and test sets for the \textbf{anaphora} task, across prompt configurations. Validation uses the first 100 couples and reproduces the scores from Table~\ref{tab:ana-accuracy-inconsistency-grid} above; test uses the held-out 100 couples. Each couple is presented in both option orders (correct option first/second), and scores are averaged across the two presentations.}
\label{tab:anaphora-test-results}
\end{table}
}

\FloatBarrier 

\begin{figure}[ht]
  \centering
  \includegraphics[width=.80\linewidth]{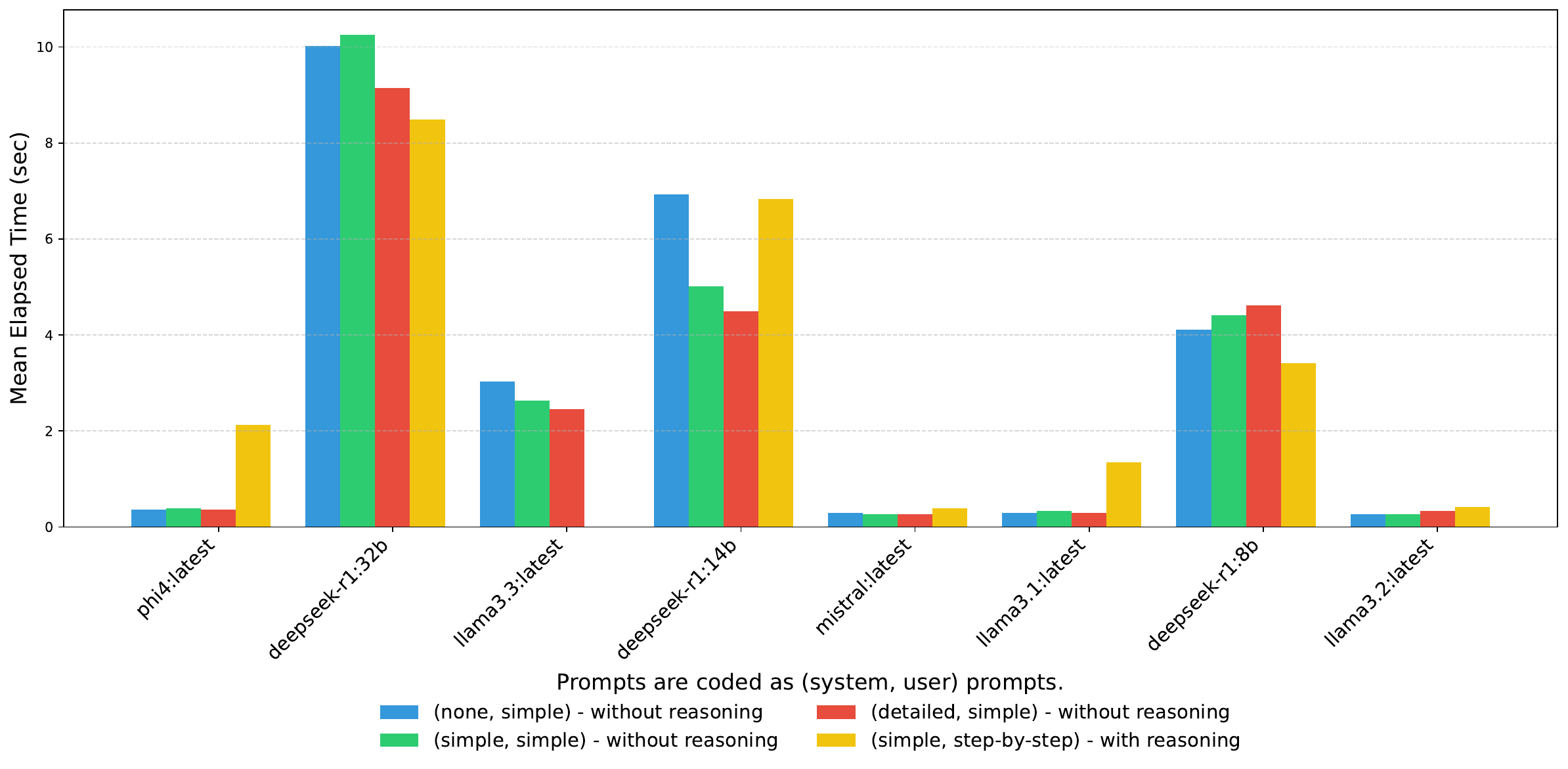}
  \caption{Average response time per prompt for open-source models on \textbf{anaphora}. Larger models tend to be slower; complex prompts increase latency.}
  \label{fig:app-ana-latency-ollama}
\end{figure}

\begin{figure}[ht]
  \centering
  \includegraphics[width=.80\linewidth]{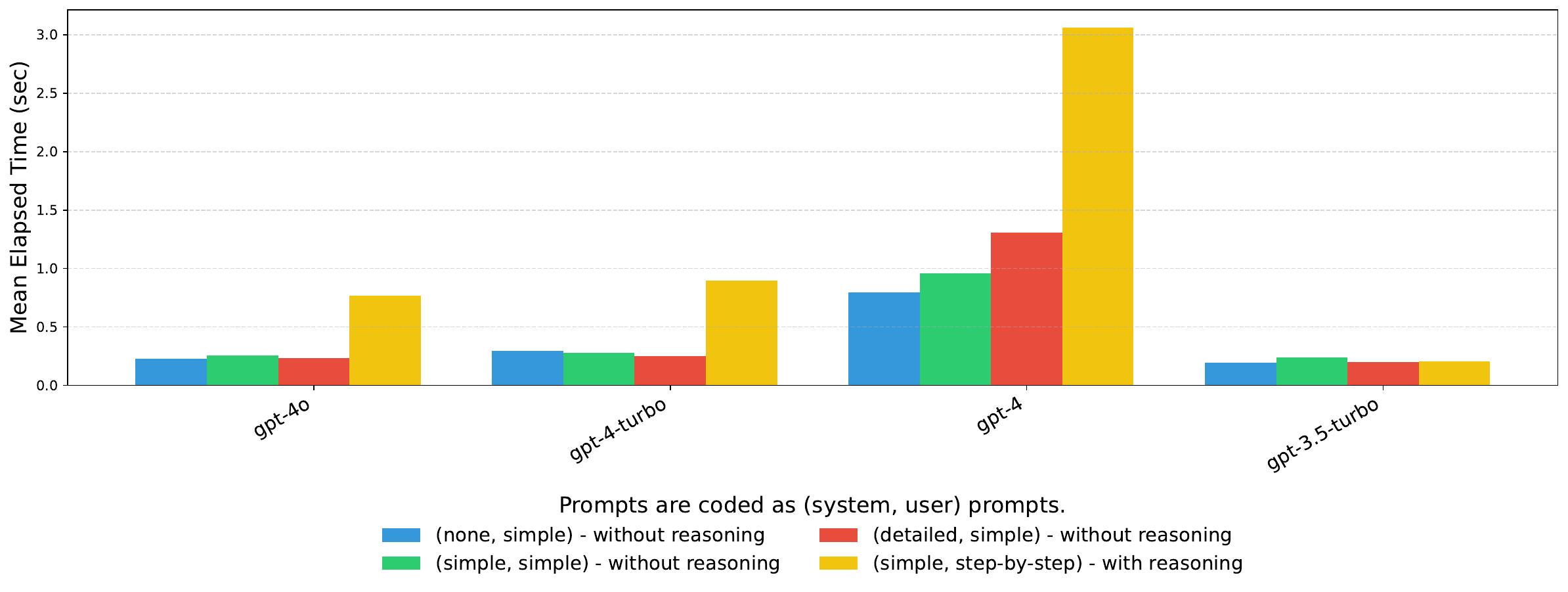}
  \caption{Average response time per prompt for OpenAI models on \textbf{anaphora}. GPT-4 is slowest overall, especially with the reasoning prompt.}
  \label{fig:app-ana-latency-openai}
\end{figure}

\begin{figure}[ht]
  \centering
  \includegraphics[width=.80\textwidth]{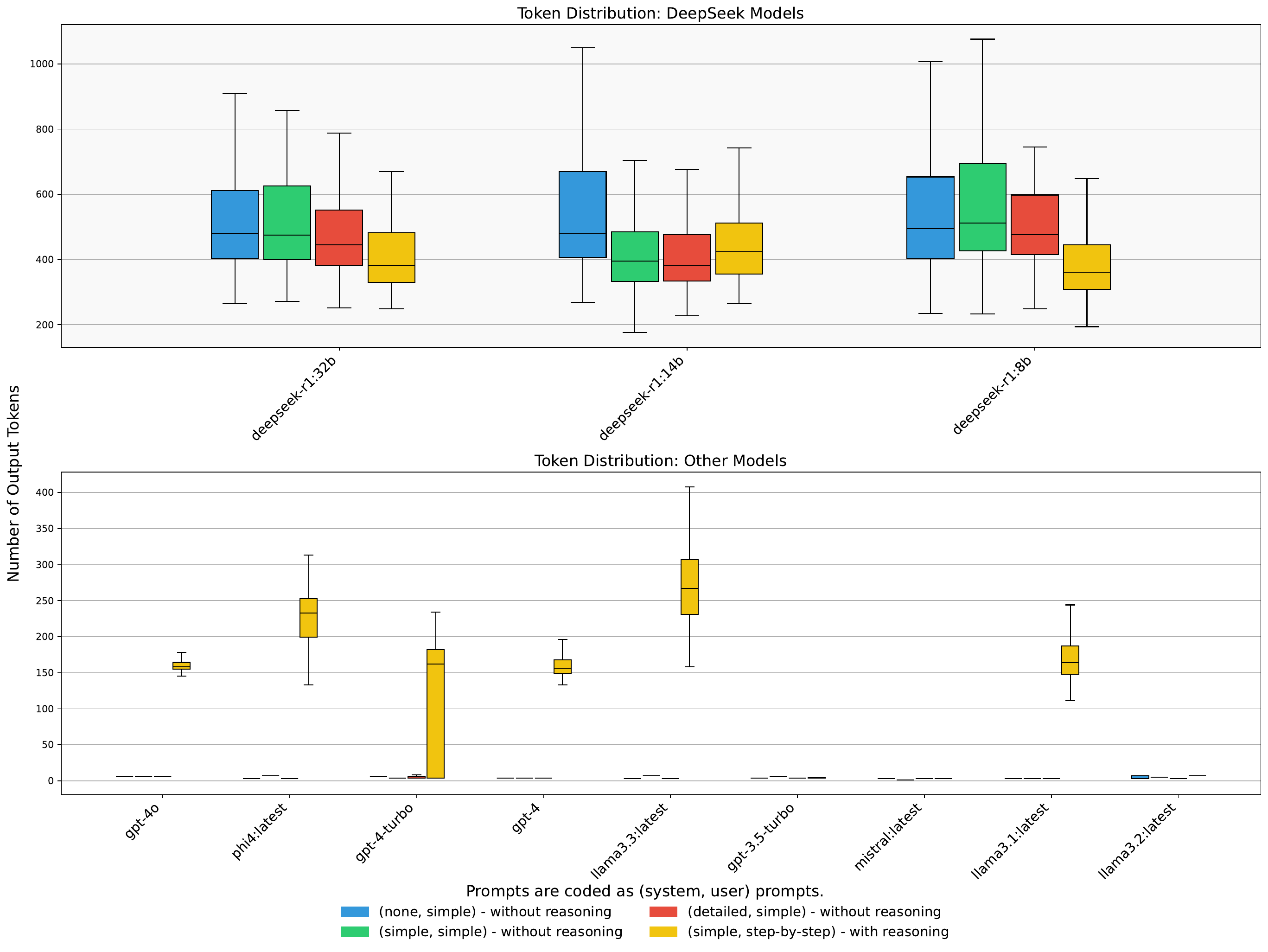}
  \caption{Distribution of output token lengths by model and prompt on \textbf{anaphora}. Top: DeepSeek models (consistently longer outputs). Bottom: all other models. Reasoning prompts generally increase length, though the effect varies by model.}
  \label{fig:app-ana-output-tokens}
\end{figure}

\begin{figure}[ht]
  \centering
  \includegraphics[width=.80\linewidth]{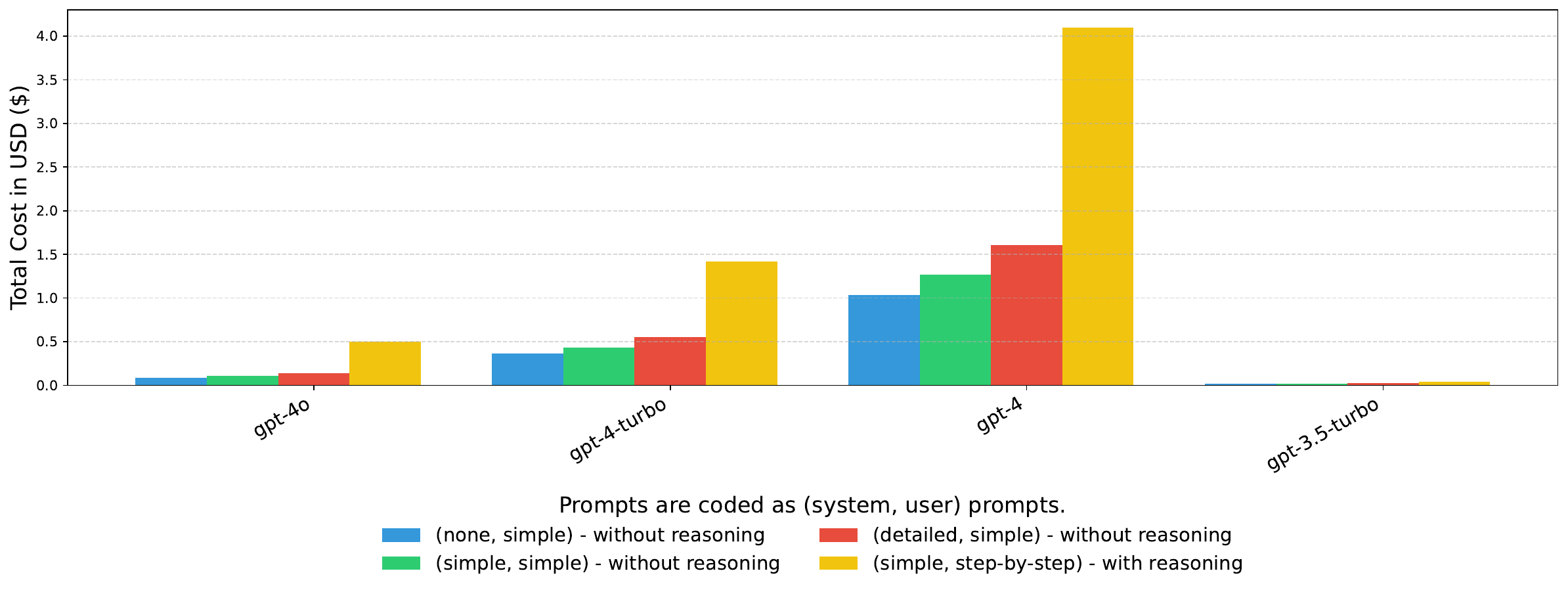}
  \caption{Total API cost by prompt configuration for OpenAI models on \textbf{anaphora}; reasoning prompts consume more tokens and cost more. Experiments were run in spring 2025, so costs reflect that period's pricing and our token usage.}
  \label{fig:app-ana-cost}
\end{figure}

\FloatBarrier 
\subsection{Results: Lexical Choice}

Below are the results on the \textbf{contrastive lexical choice benchmark}, for each LLM and prompt type, in terms of accuracy and inconsistency scores, elapsed time and cost of LLM.

{
\setlength{\abovecaptionskip}{8pt}
\begin{table}[ht]
\centering
\small
\setlength{\tabcolsep}{5pt}\renewcommand{\arraystretch}{1.05}
\resizebox{0.95\textwidth}{!}{%
\begin{tabular}{|l|c|c|c|c|}
\hline
 & \multicolumn{4}{c|}{\textbf{System and user prompts}} \\
\textbf{Model} & \textbf{None \& simple} & \textbf{Both simple} & \textbf{Detailed \& simple)} & \textbf{Simple \& step-by-step} \\
\hline
gpt-4o            & 0.86 / 0.04 & 0.88 / 0.04 & 0.87 / 0.03 & 0.96 / 0.02 \\
\hline
gpt-4             & 0.89 / 0.03 & 0.90 / 0.04 & 0.89 / 0.04 & 0.92 / 0.02 \\
\hline
gpt-4-turbo       & 0.87 / 0.01 & 0.84 / 0.02 & 0.84 / 0.03 & 0.88 / 0.00 \\
\hline
LLaMA 3.3 (70B)   & 0.86 / 0.04 & 0.86 / 0.02 & 0.86 / 0.00 & 0.82 / 0.03 \\
\hline
Phi-4 (14B)       & 0.83 / 0.00 & 0.83 / 0.01 & 0.84 / 0.07 & 0.80 / 0.05 \\
\hline
gpt-3.5-turbo     & 0.80 / 0.05 & 0.83 / 0.05 & 0.84 / 0.04 & 0.80 / 0.08 \\
\hline
DeepSeek-R1 (14B) & 0.76 / 0.05 & 0.76 / 0.01 & 0.78 / 0.03 & 0.79 / 0.03 \\
\hline
Mistral (7B)      & 0.58 / 0.69 & 0.50 / 1.00 & 0.64 / 0.46 & 0.76 / 0.19 \\
\hline
DeepSeek-R1 (32B) & 0.75 / 0.04 & 0.75 / 0.00 & 0.76 / 0.08 & 0.74 / 0.02 \\
\hline
DeepSeek-R1 (8B)  & 0.67 / 0.07 & 0.60 / 0.11 & 0.62 / 0.12 & 0.66 / 0.02 \\
\hline
LLaMA 3.1 (8B)    & 0.76 / 0.13 & 0.78 / 0.01 & 0.77 / 0.05 & 0.64 / 0.12 \\
\hline
LLaMA 3.2 (3B)    & 0.66 / 0.32 & 0.58 / 0.57 & 0.68 / 0.24 & 0.52 / 0.90 \\
\hline
\end{tabular}}
\caption{Accuracy (\(\uparrow\)) and inconsistency (\(\downarrow\)) across models and prompts on the \textbf{contrastive lexical  choice} validation set (separated by `/' in each cell). The systems are ranked by decreasing accuracy of the best overall prompt, which is the fourth one.}
\label{tab:accuracy-inconsistency-lexical}
\end{table}
}

{
\setlength{\abovecaptionskip}{8pt}
\begin{table}[ht]
\centering
\small
\setlength{\tabcolsep}{6pt}\renewcommand{\arraystretch}{1.05}
\resizebox{0.95\textwidth}{!}{%
\begin{tabular}{|l|c|c|c|c|}
\hline
 & \multicolumn{4}{c|}{\textbf{System and user prompts}} \\
\textbf{Model} & \textbf{None \& simple} & \textbf{Both simple} & \textbf{Detailed \& simple)} & \textbf{Simple \& step-by-step} \\
\hline
LLaMA 3.3 (70B)   & 2.49 & 2.79 & 2.16 & 52.53 \\
\hline
Phi-4 (14B)       & 0.33 & 0.36 & 0.30 & 1.95 \\
\hline
DeepSeek-R1 (14B) & 4.53 & 3.51 & 3.21 & 4.47 \\
\hline
Mistral (7B)      & 0.27 & 0.24 & 0.27 & 0.33 \\
\hline
DeepSeek-R1 (32B) & 8.55 & 7.92 & 6.27 & 7.29 \\
\hline
DeepSeek-R1 (8B)  & 2.97 & 3.09 & 2.82 & 2.61 \\
\hline
LLaMA 3.1 (8B)    & 0.30 & 0.27 & 0.30 & 1.35 \\
\hline
LLaMA 3.2 (3B)    & 0.24 & 0.24 & 0.24 & 0.39 \\
\hline\hline
gpt-4o            & 0.408 & 0.498 & 0.444 & 1.788 \\
\hline
gpt-4             & 0.504 & 0.480 & 0.480 & 3.432 \\
\hline
gpt-4-turbo       & 0.552 & 0.564 & 0.576 & 1.404 \\
\hline
gpt-3.5-turbo     & 0.348 & 0.348 & 0.360 & 0.396 \\
\hline
\end{tabular}}
\caption{Mean elapsed time per prompt (seconds) across prompt configurations on the \textbf{lexical choice} validation set. Open-source (Ollama) models listed first, followed by OpenAI models.}
\label{tab:latency-all-lexical}
\end{table}
}

{
\setlength{\abovecaptionskip}{8pt}
\begin{table}[ht]
\centering
\small
\setlength{\tabcolsep}{6pt}\renewcommand{\arraystretch}{1.05}
\resizebox{0.95\textwidth}{!}{
\begin{tabular}{|l|c|c|c|c|}
\hline
 & \multicolumn{4}{c|}{\textbf{System and user prompts}} \\
\textbf{Model} & \textbf{None \& simple} & \textbf{Both simple} & \textbf{Detailed \& simple)} & \textbf{Simple \& step-by-step} \\
\hline
gpt-4o         & 0.083888 & 0.103422 & 0.131416 & 0.495446 \\
\hline
gpt-4          & 0.952440 & 1.183200 & 1.519200 & 3.696600 \\
\hline
gpt-4-turbo    & 0.335800 & 0.402520 & 0.524720 & 1.063320 \\
\hline
gpt-3.5-turbo  & 0.016230 & 0.020799 & 0.025751 & 0.036889 \\
\hline
\end{tabular}}
\caption{Total API cost (USD) by prompt configuration for OpenAI models on the \textbf{lexical choice} validation set. Experiments were run in summer~2025; reported costs reflect that period's pricing and our token usage.}
\label{tab:cost-all-lexical}
\end{table}
}

{%
\setlength{\abovecaptionskip}{8pt}
\begin{table}[ht]
\centering
\small
\setlength{\tabcolsep}{5pt}\renewcommand{\arraystretch}{1.05}
\resizebox{0.95\textwidth}{!}{%
\begin{tabular}{|l|l|l|c|c|c|c|}
\hline
\textbf{Model} & \textbf{System} & \textbf{User} & \textbf{Val. Acc.} & \textbf{Val. Inc.} & \textbf{Test Acc.} & \textbf{Test Inc.} \\
\hline
gpt-4o              & none     & simple        & 0.86 & 0.04 & 0.94 & 0.03 \\
gpt-4o              & simple   & simple        & 0.88 & 0.04 & 0.94 & 0.04 \\
gpt-4o              & detailed & simple        & 0.87 & 0.03 & 0.94 & 0.02 \\
gpt-4o              & simple   & step-by-step  & 0.96 & 0.02 & 0.96 & 0.01 \\
\hline
gpt-4-turbo         & none     & simple        & 0.87 & 0.01 & 0.92 & 0.03 \\
gpt-4-turbo         & simple   & simple        & 0.84 & 0.02 & 0.92 & 0.03 \\
gpt-4-turbo         & detailed & simple        & 0.84 & 0.03 & 0.92 & 0.04 \\
gpt-4-turbo         & simple   & step-by-step  & 0.88 & 0.00 & 0.94 & 0.02 \\
\hline
Phi-4 (14B)         & none     & simple        & 0.83 & 0.00 & 0.89 & 0.01 \\
Phi-4 (14B)         & simple   & simple        & 0.83 & 0.01 & 0.91 & 0.02 \\
Phi-4 (14B)         & detailed & simple        & 0.84 & 0.07 & 0.90 & 0.03 \\
Phi-4 (14B)         & simple   & step-by-step  & 0.80 & 0.05 & 0.92 & 0.02 \\
\hline
gpt-4               & none     & simple        & 0.89 & 0.03 & 0.94 & 0.01 \\
gpt-4               & simple   & simple        & 0.90 & 0.04 & 0.92 & 0.02 \\
gpt-4               & detailed & simple        & 0.89 & 0.04 & 0.92 & 0.03 \\
gpt-4               & simple   & step-by-step  & 0.92 & 0.02 & 0.91 & 0.02 \\
\hline
LLaMA 3.3 (70B)     & none     & simple        & 0.86 & 0.04 & 0.88 & 0.00 \\
LLaMA 3.3 (70B)     & simple   & simple        & 0.86 & 0.02 & 0.87 & 0.00 \\
LLaMA 3.3 (70B)     & detailed & simple        & 0.86 & 0.00 & 0.86 & 0.04 \\
LLaMA 3.3 (70B)     & simple   & step-by-step  & 0.82 & 0.03 & 0.90 & 0.04 \\
\hline
gpt-3.5-turbo       & none    & simple        & 0.80 & 0.05 & 0.84 & 0.04 \\
gpt-3.5-turbo       & simple   & simple        & 0.83 & 0.05 & 0.86 & 0.02 \\
gpt-3.5-turbo       & detailed & simple        & 0.84 & 0.04 & 0.86 & 0.05 \\
gpt-3.5-turbo       & simple   & step-by-step  & 0.80 & 0.08 & 0.84 & 0.00 \\
\hline
DeepSeek-R1 (14B)   & none     & simple        & 0.76 & 0.05 & 0.86 & 0.04 \\
DeepSeek-R1 (14B)   & simple   & simple        & 0.76 & 0.01 & 0.88 & 0.03 \\
DeepSeek-R1 (14B)   & detailed & simple        & 0.78 & 0.03 & 0.84 & 0.07 \\
DeepSeek-R1 (14B)   & simple   & step-by-step  & 0.79 & 0.03 & 0.78 & 0.01 \\
\hline
\end{tabular}}
\caption{Comparison of accuracy (Acc \(\uparrow\)) and inconsistency (Inc \(\downarrow\)) on validation and test sets for the \textbf{lexical choice} task, across prompt configurations. Validation uses the first 100 couples and reproduces the scores from Table~\ref{tab:accuracy-inconsistency-lexical} above; test uses the held-out 100 couples. Each couple is presented in both option orders (correct option first/second), and scores are averaged across the two presentations.}
\label{tab:lexical-choice-test-results}
\end{table}
}

\FloatBarrier
\subsubsection{Graphical Representation of Scores}


\begin{figure}[ht]
  \centering
  \includegraphics[width=.80\textwidth]{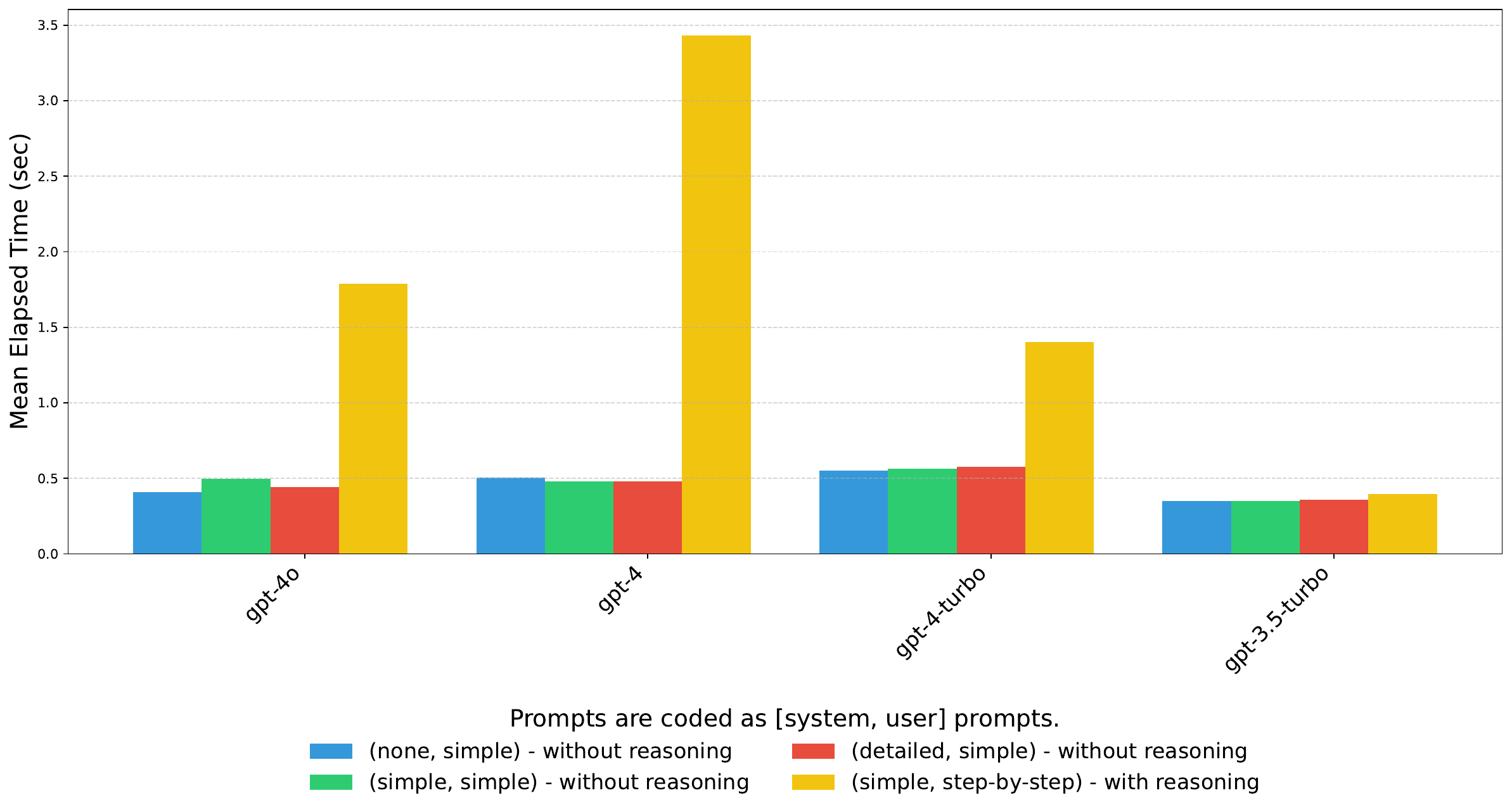}
  \caption{Average response time per prompt for OpenAI models on the \textbf{lexical choice} validation set. GPT-4 has the highest latency under reasoning prompts.}
  \label{fig:app-lex-latency-openai}
\end{figure}

\begin{figure}[ht]
  \centering
  \includegraphics[width=.80\textwidth]{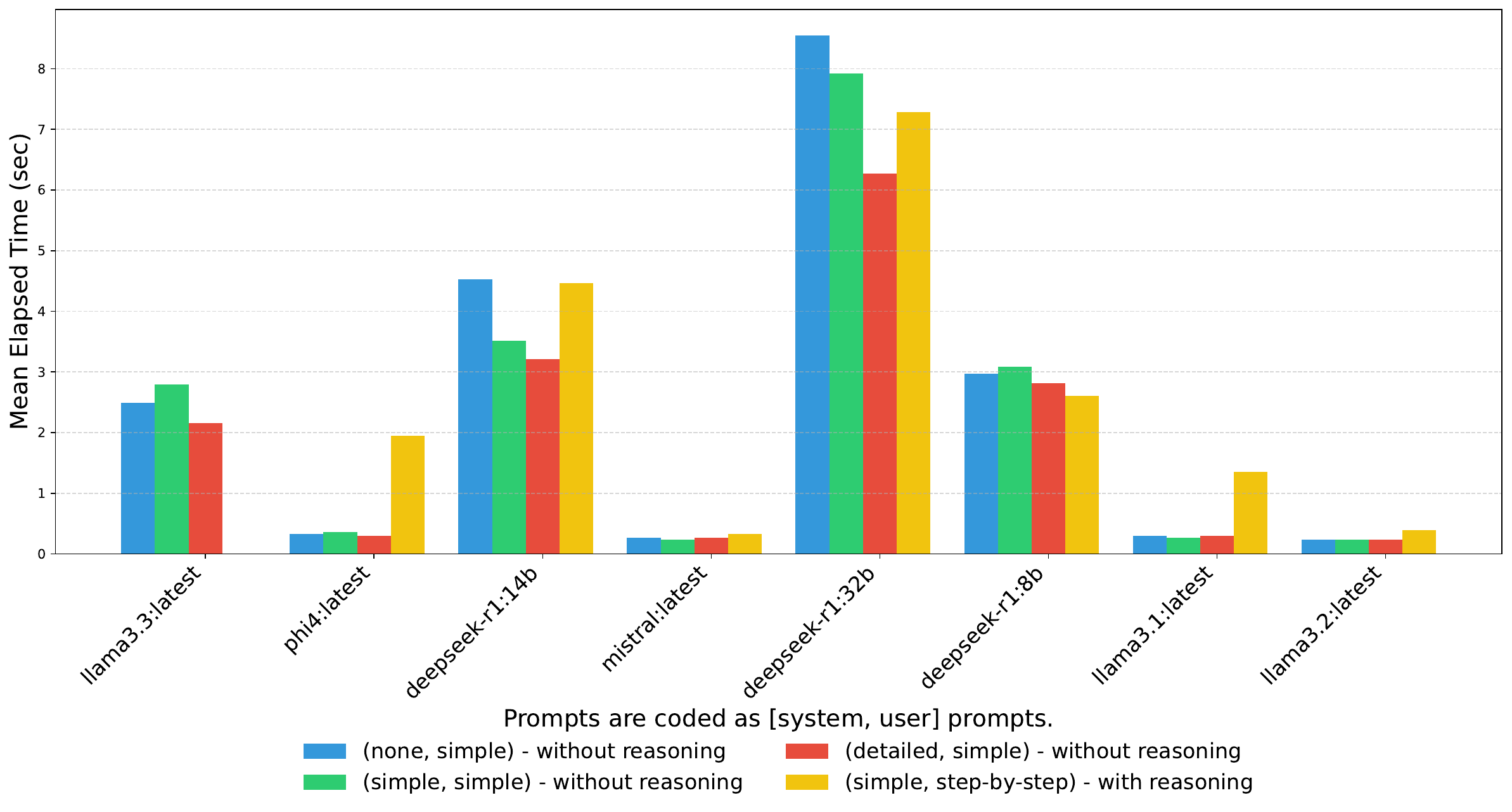}
  \caption{Average response time per prompt for open-source models on the \textbf{lexical choice} validation set. Phi-4 is fastest; DeepSeek and LLaMA 3.3 are notably slower.}
  \label{fig:app-lex-latency-ollama}
\end{figure}

\begin{figure}[ht]
  \centering
  \includegraphics[width=.80\textwidth]{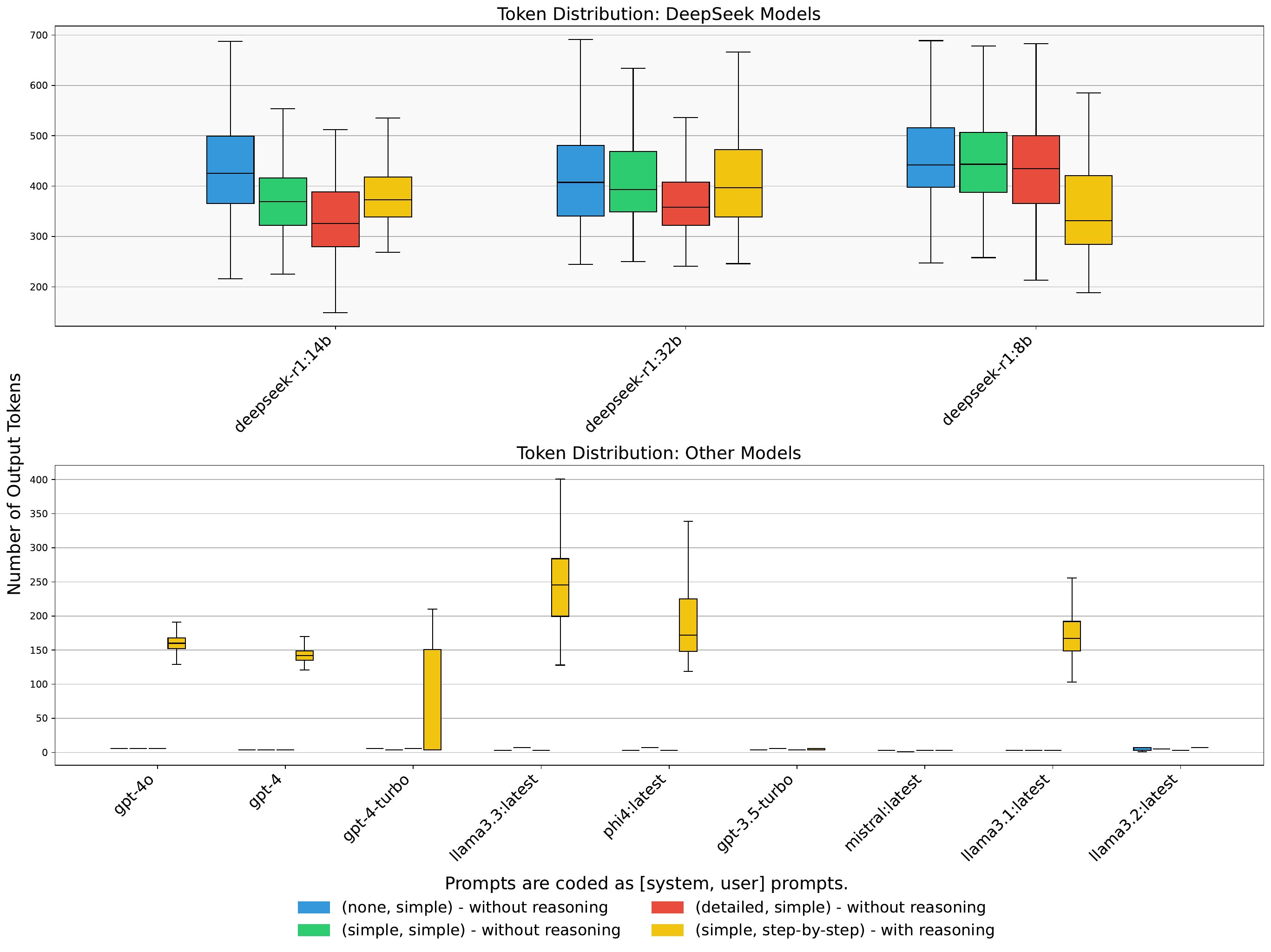}
  \caption{Distribution of output token lengths by model and prompt on the \textbf{lexical choice} validation set. DeepSeek models produce notably longer outputs; most other models remain concise under the prompts.}
  \label{fig:app-lex-output-tokens}
\end{figure}

\begin{figure}[ht]
  \centering
  \includegraphics[width=.80\textwidth]{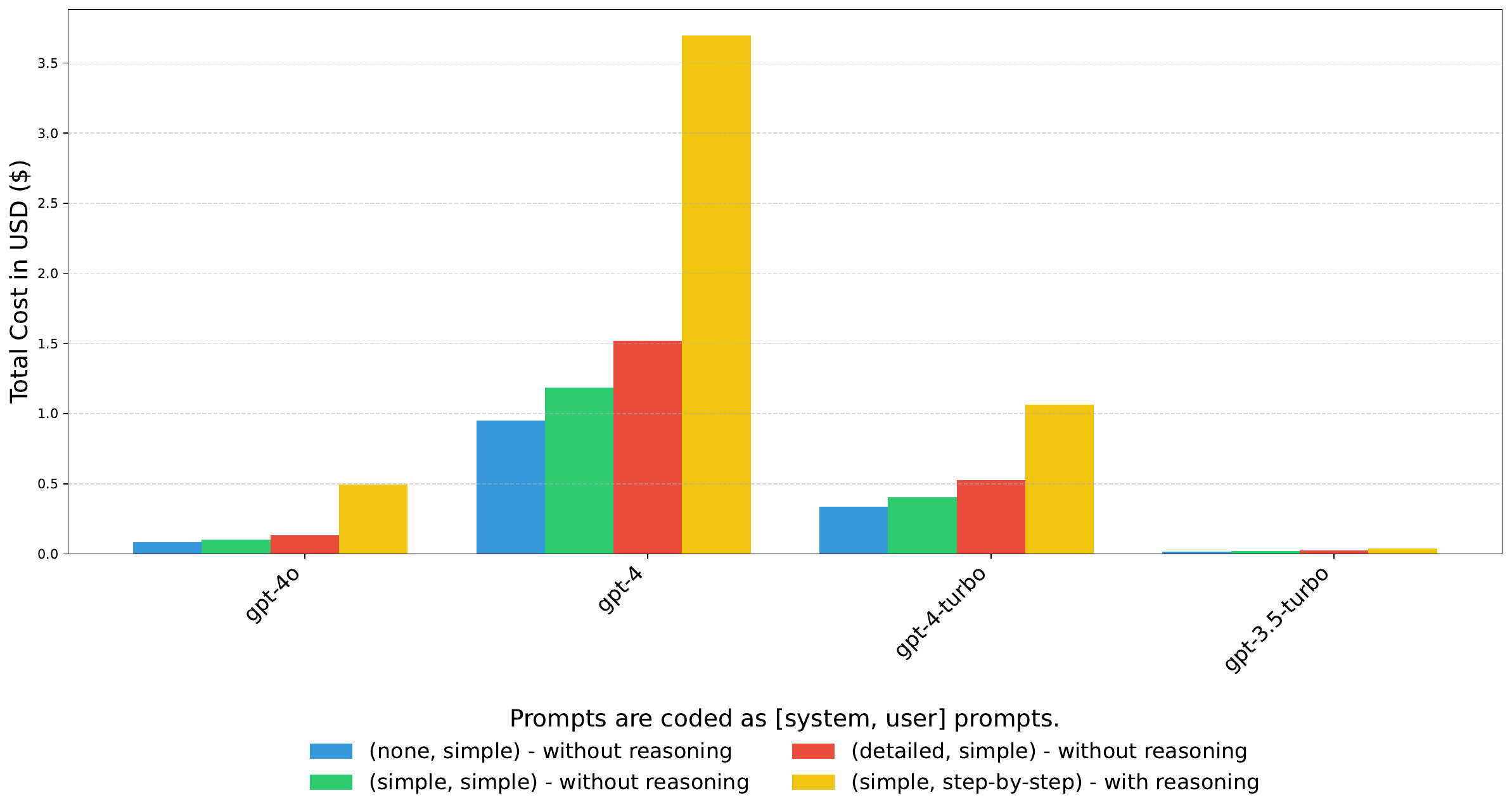}
  \caption{Total API cost by prompt configuration for OpenAI models on the \textbf{lexical choice} validation set; reasoning prompts increase token usage and cost, especially for GPT-4. Experiments were run in spring~2025, so costs reflect that period's pricing and our token usage.}
  
  \label{fig:app-lex-cost}
\end{figure}

\FloatBarrier

\section{Generative Task}

\subsection{Prompt Variants}
\label{sec:prompt-variants-translation}

This section presents the full text of the prompts used in our \textit{generative} evaluation (translation task).  
Similarly to the contrastive prompts, each configuration has a \textit{system prompt} and a \textit{user prompt}.
and the placeholders appear in curly braces: \texttt{\{context\_en\}}, \texttt{\{source\_sentence\}}, \texttt{\{context\_fr\}}, \texttt{\{option1\}}, \texttt{\{option2\}}.
In the generative setting, the model must produce the French translation of the second sentence, with output format rules depending on the reasoning prompt used.

\paragraph{Prompt without reasoning -- identical for anaphora and lexical choice}\mbox{}\\[0.5em]
\indent\textit{System prompt}\\
{\ttfamily\footnotesize
You are a professional translator. Your task is to translate short English texts into French.\\
\\
\#\#\# Instructions:\\
- You will receive two English sentences: a context sentence and a sentence to translate.\\
- You will also be given the French translation of the context sentence.\\
- Translate ONLY the second English sentence into French.\\
- Return ONLY the French translation of the second sentence.\\
- Do NOT include any explanation or additional text.
}

\textit{User prompt}\\
{\ttfamily\footnotesize
Here is a short text with two sentences in English:\\
\texttt{\{context\_en\}}\\
\texttt{\{source\_sentence\}}\\
Here is the French translation of the first sentence:\\
\texttt{\{context\_fr\}}\\
Please translate the second sentence into French.
}

\vspace{0.5em}

\paragraph{Structured reasoning (XML-style) -- Anaphora}\mbox{}\\[0.5em]
\indent\textit{System prompt}\\
{\ttfamily\footnotesize
You are a professional translator. Your task is to translate short English texts into French.\\
\\
\#\#\# Instructions:\\
- You will receive two English sentences: a context sentence and a sentence to translate.\\
- You will also be given the French translation of the first sentence.\\
- Translate the second English sentence into French.\\
- To achieve this, think step by step to resolve pronouns and references correctly using the context:\\
\ \ \ 1.\ Identify pronouns/references in the second sentence.\\
\ \ \ 2.\ Find their referent in the first sentence.\\
\ \ \ 3.\ Check how that referent is translated in the French context.\\
\ \ \ 4.\ Choose the correct French pronoun/reference.\\
- Finally, output only this XML (valid single root):\\
\ \ \ \textless result\textgreater\\
\ \ \ \ \ \ \textless reasoning\textgreater\textless/reasoning\textgreater\\
\ \ \ \ \ \ \textless answer\textgreater\textless/answer\textgreater\\
\ \ \ \textless/result\textgreater
}

\textit{User prompt}\\
{\ttfamily\footnotesize
Here is a short English passage:\\
\texttt{\{context\_en\}}\\
\texttt{\{source\_sentence\}}\\
Here is the French translation of the first sentence:\\
\texttt{\{context\_fr\}}\\
Please translate the second sentence into French, following the instructions.\\
Output only the XML format specified.
}

\vspace{0.5em}

\paragraph{(Structured reasoning (XML-style) -- Lexical Choice}\mbox{}\\[0.5em]
\indent\textit{System prompt}\\
{\ttfamily\footnotesize
You are a professional translator. Your task is to translate short English texts into French.\\
\\
\#\#\# Instructions:\\
- You will receive two English sentences: a first sentence and a second sentence.\\
- You will also receive the French translation of the first sentence.\\
- Translate the second English sentence into French.\\
\\
To ensure lexical consistency:\\
1.\ Identify any key word or expression in the second sentence that could be translated in more than one way into French.\\
2.\ Check whether the French translation of the first sentence already provides a preferred translation for that word or expression,\\
\ \ \ and use the same choice if appropriate for your translation of the second sentence.\\
\\
- Finally, output \emph{only} this XML (valid single root):\\
\ \ \ \textless result\textgreater\\
\ \ \ \ \ \ \textless reasoning\textgreater\textless/reasoning\textgreater\\
\ \ \ \ \ \ \textless answer\textgreater\textless/answer\textgreater\\
\ \ \ \textless/result\textgreater
}

\textit{User prompt}\\
{\ttfamily\footnotesize
Here is a short English text with two sentences:\\
\texttt{\{context\_en\}}\\
\texttt{\{source\_sentence\}}\\
Here is the French translation of the first sentence:\\
\texttt{\{context\_fr\}}\\
Please translate the second sentence into French, following the instructions in the system prompt.\\
Output only the XML format specified.
}

\vspace{0.5em}

\FloatBarrier
\subsection{Results: Anaphora}
\FloatBarrier

{%
\setlength{\abovecaptionskip}{8pt}
\begin{table}[ht]
\centering
\small
\setlength{\tabcolsep}{5pt}\renewcommand{\arraystretch}{1.05}
\resizebox{0.95\textwidth}{!}{%
\begin{tabular}{|l|ccc|ccc|ccc|ccc|}
\hline
\textbf{Model} &
\multicolumn{3}{c|}{\textbf{BLEU} \(\uparrow\)} &
\multicolumn{3}{c|}{\textbf{chrF} \(\uparrow\)} &
\multicolumn{3}{c|}{\textbf{BERTScore (2nd)} \(\uparrow\)} &
\multicolumn{3}{c|}{\textbf{COMET (2nd)} \(\uparrow\)} \\
 & \textbf{w/o} & \textbf{w/} & \(\boldsymbol{\Delta}\) & \textbf{w/o} & \textbf{w/} & \(\boldsymbol{\Delta}\) & \textbf{w/o} & \textbf{w/} & \(\boldsymbol{\Delta}\) & \textbf{w/o} & \textbf{w/} & \(\boldsymbol{\Delta}\) \\
\hline
gpt-4            & 49.08 & 52.86 & +3.78 & 69.58 & 71.35 & +1.77 & 0.9242 & 0.9275 & +0.0033 & 0.9161 & 0.9167 & +0.0006 \\
gpt-4o           & 49.06 & 54.41 & +5.35 & 69.53 & 71.49 & +1.96 & 0.9234 & 0.9250 & +0.0016 & 0.9101 & 0.9118 & +0.0017 \\
gpt-4-turbo      & 45.00 & 49.17 & +4.17 & 66.51 & 67.41 & +0.90 & 0.9163 & 0.8990 & -0.0173 & 0.9114 & 0.9030 & -0.0084 \\
gpt-3.5-turbo    & 43.97 & 49.19 & +5.22 & 65.66 & 67.81 & +2.15 & 0.9125 & 0.9142 & +0.0017 & 0.9116 & 0.9036 & -0.0080 \\
LLaMA 3.3        & 43.90 & 46.58 & +2.68 & 66.30 & 67.38 & +1.08 & 0.9179 & 0.9133 & -0.0046 & 0.9019 & 0.8976 & -0.0043 \\
Phi-4            & 43.43 & 49.01 & +5.58 & 63.89 & 68.28 & +4.39 & 0.9084 & 0.9180 & +0.0096 & 0.8782 & 0.9052 & +0.0270 \\
DeepSeek-R1 32B  & 35.02 & 39.36 & +4.34 & 58.85 & 58.81 & -0.04 & 0.8930 & 0.8379 & -0.0551 & 0.8736 & 0.8450 & -0.0286 \\
LLaMA 3.1        & 34.13 & 30.35 & -3.78 & 57.76 & 54.41 & -3.35 & 0.8891 & 0.8678 & -0.0213 & 0.8594 & 0.8464 & -0.0130 \\
DeepSeek-R1 14B  & 34.04 & 32.98 & -1.06 & 58.20 & 53.29 & -4.91 & 0.8855 & 0.7928 & -0.0927 & 0.8551 & 0.7904 & -0.0647 \\
Mistral          & 26.69 & 26.66 & -0.03 & 50.59 & 49.79 & -0.80 & 0.8553 & 0.8469 & -0.0084 & 0.8229 & 0.8047 & -0.0182 \\
LLaMA 3.2        & 26.67 & 23.19 & -3.48 & 51.24 & 47.41 & -3.83 & 0.8667 & 0.8277 & -0.0390 & 0.8163 & 0.7584 & -0.0579 \\
DeepSeek-R1 8B   & 23.74 & 21.26 & -2.48 & 49.96 & 45.21 & -4.75 & 0.8586 & 0.7678 & -0.0908 & 0.8010 & 0.7712 & -0.0298 \\
\hline
\end{tabular}}
\caption{Translation task on the \textbf{anaphora benchmark} using a structured (XML-style) reasoning prompt. Scores are aggregated over 200 examples (validation+test). Columns show performance \textit{without} reasoning (w/o), \textit{with} reasoning (w/), and the difference \(\Delta = \text{w/} - \text{w/o}\).}
\label{tab:anaphora-generative-newprompt}
\end{table}
}

\FloatBarrier 

\begin{figure}[ht]
  \centering
  \includegraphics[width=.80\linewidth]{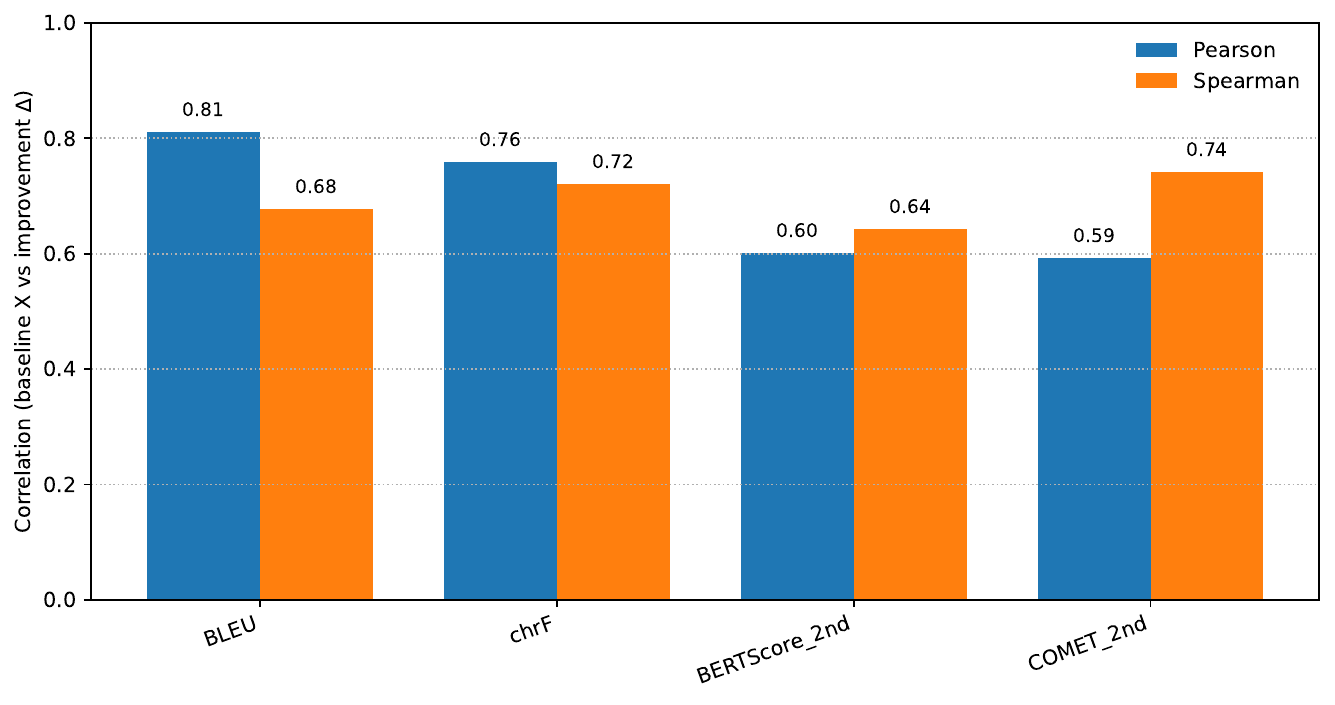}
  \caption{Correlation between baseline (no-reasoning) scores and gains \(\Delta\) under the structured (XML-style) reasoning prompt. Bars show Pearson and Spearman coefficients for BLEU, chrF, BERTScore, and COMET. Positive correlations across metrics support the “wise get wiser” effect: higher-performing models tend to benefit more from reasoning.}
  \label{fig:app-gen-ana-wise-bar}
\end{figure}

\begin{figure}[ht]
  \centering
  \includegraphics[width=0.95\linewidth]{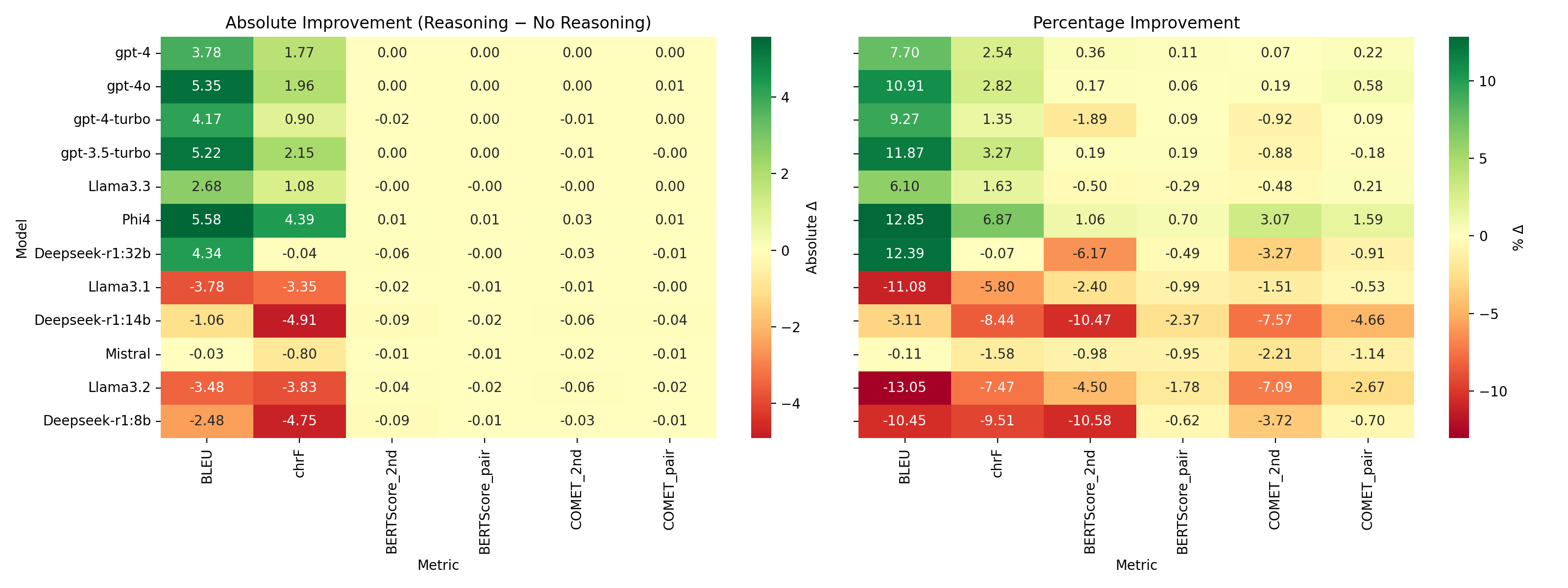}
  \caption{Heatmaps of gains from the structured (XML-style) reasoning prompt. Left: absolute changes for BLEU, chrF, BERTScore, and COMET; right: the same changes as percentages relative to the no-reasoning baseline. Positive cells indicate improvements; negative cells indicate declines, allowing a quick scan of which models/metrics benefit from reasoning.}
  \label{fig:app-gen-ana-improv-heatmaps}
\end{figure}

\begin{figure}[ht]
  \centering
  \includegraphics[width=.8\linewidth]{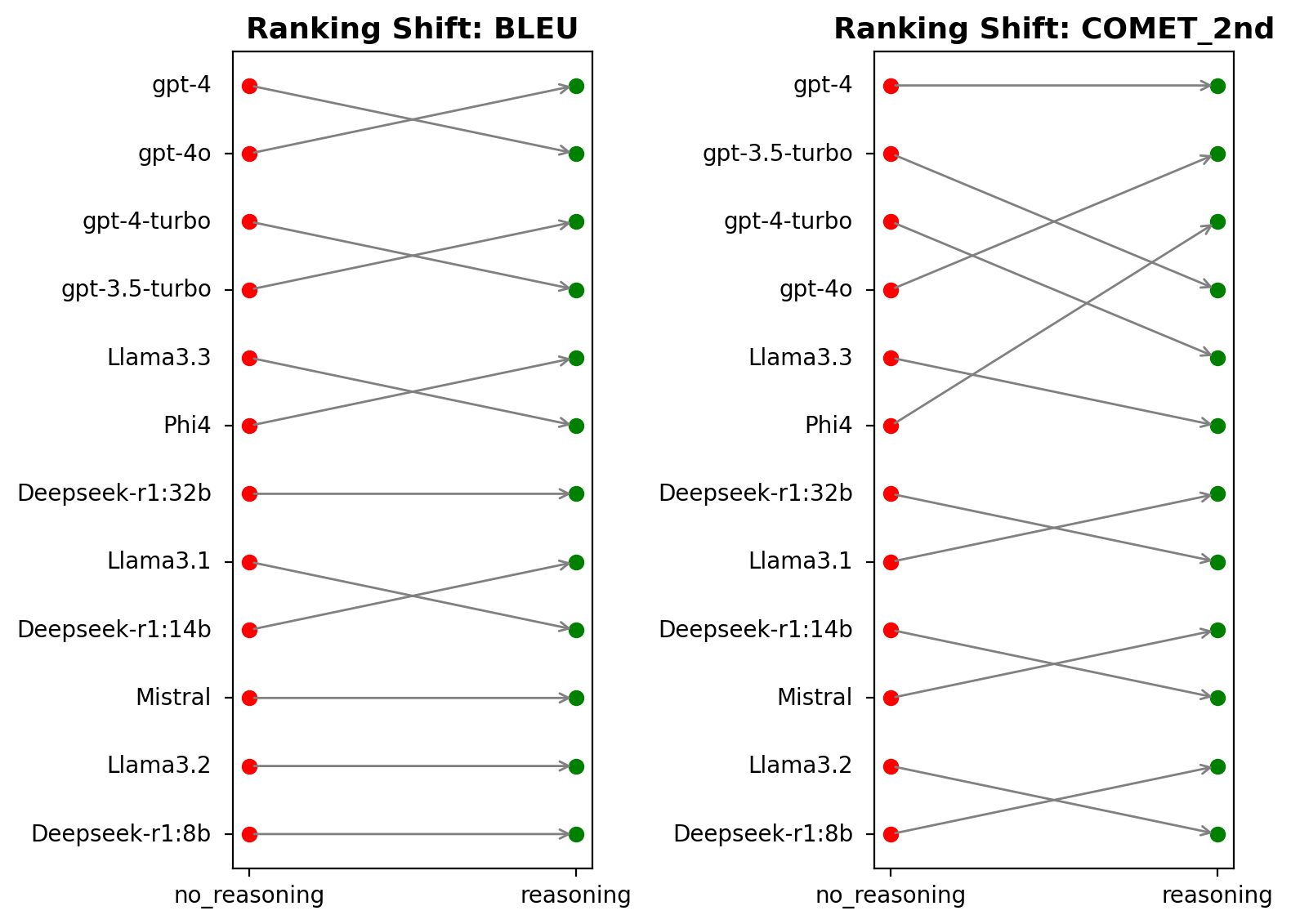}
  \caption{Ranking shifts from no reasoning (left) to structured reasoning (right) for BLEU and COMET. Each line traces a model’s rank: upward moves indicate improved ranking; crossings show notable re-orderings. This highlights which systems improve their rank relative to others when reasoning is enabled.}
  \label{fig:app-gen-ana-ranking-blue-comet}
\end{figure}

\begin{figure}[ht]
  \centering
  \includegraphics[width=.80\linewidth]{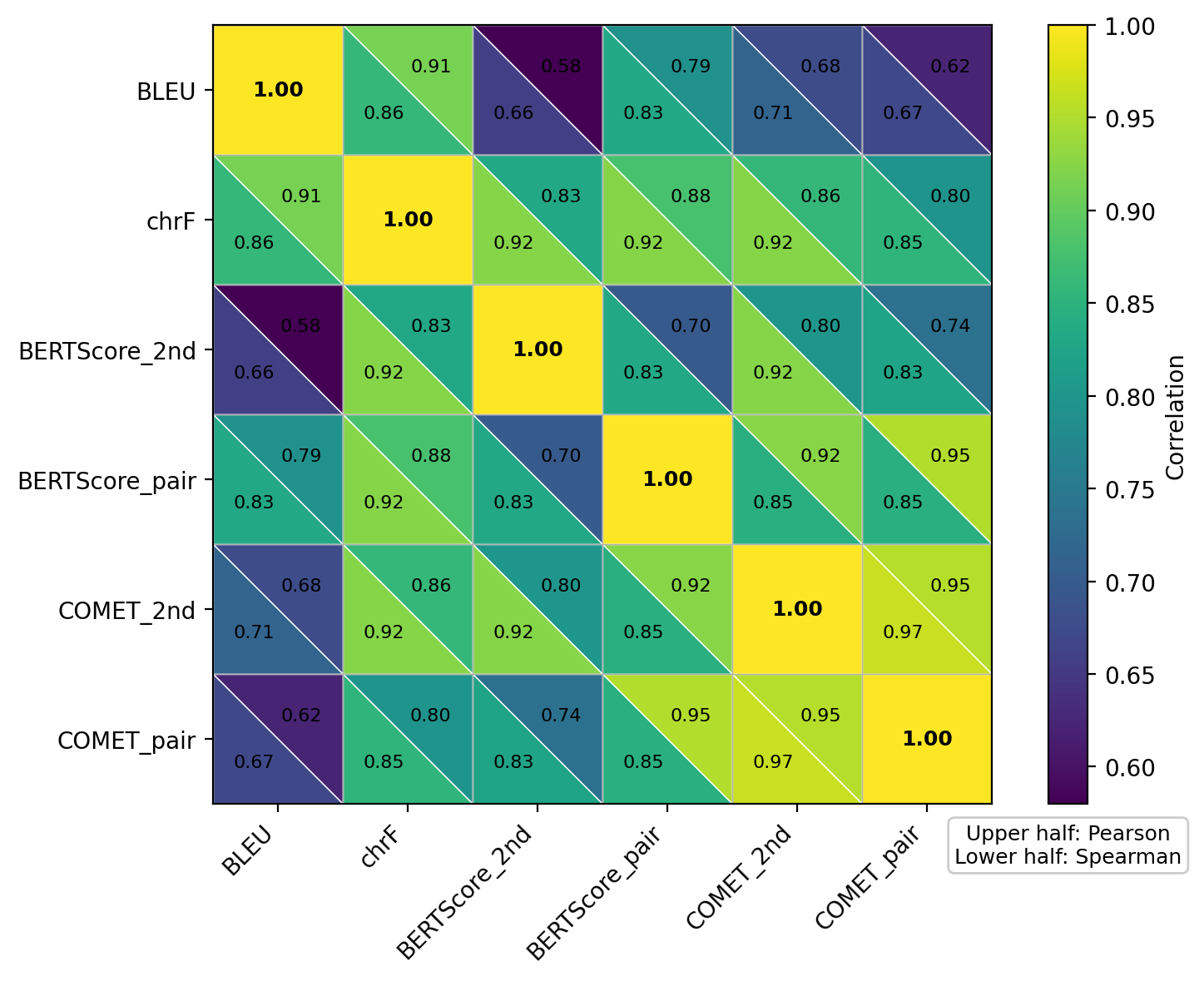}
  \caption{Correlation matrix of improvements \(\Delta\) (reasoning \( - \) no\_reasoning) under the structured (XML-style) prompt across twelve models. Each cell reports Pearson (upper triangle) and Spearman (lower triangle); colors are scaled to the observed min--max range for contrast. Strong positive cells indicate that models improving on one metric tend to improve on others as well.}
  \label{fig:app-gen-ana-corr}
\end{figure}

\FloatBarrier 
\subsection{Results: Lexical Choice}

{%
\setlength{\belowcaptionskip}{8pt}
\begin{table}[ht]
\centering
\small
\setlength{\tabcolsep}{5pt}\renewcommand{\arraystretch}{1.05}
\resizebox{0.95\textwidth}{!}{%
\begin{tabular}{|l|ccc|ccc|ccc|ccc|}
\hline
\textbf{Model} &
\multicolumn{3}{c|}{\textbf{BLEU} \(\uparrow\)} &
\multicolumn{3}{c|}{\textbf{chrF} \(\uparrow\)} &
\multicolumn{3}{c|}{\textbf{BERTScore (2nd)} \(\uparrow\)} &
\multicolumn{3}{c|}{\textbf{COMET (2nd)} \(\uparrow\)} \\
 & \textbf{w/o} & \textbf{w/} & \(\boldsymbol{\Delta}\) & \textbf{w/o} & \textbf{w/} & \(\boldsymbol{\Delta}\) & \textbf{w/o} & \textbf{w/} & \(\boldsymbol{\Delta}\) & \textbf{w/o} & \textbf{w/} & \(\boldsymbol{\Delta}\) \\
\hline
gpt-4o           & 53.84 & 54.15 & +0.31 & 69.67 & 69.41 & -0.26 & 0.9223 & 0.9208 & -0.0015 & 0.8851 & 0.8797 & -0.0054 \\
gpt-4            & 51.41 & 53.99 & +2.58 & 66.63 & 68.79 & +2.16 & 0.9128 & 0.9179 & +0.0051 & 0.8643 & 0.8713 & +0.0070 \\
gpt-4-turbo      & 49.56 & 49.25 & -0.31 & 65.67 & 64.84 & -0.83 & 0.9120 & 0.8911 & -0.0209 & 0.8683 & 0.8557 & -0.0126 \\
gpt-3.5-turbo    & 47.31 & 49.42 & +2.11 & 64.88 & 65.69 & +0.81 & 0.9061 & 0.9102 & +0.0041 & 0.8619 & 0.8560 & -0.0059 \\
Llama3.3         & 46.43 & 46.58 & +0.15 & 62.47 & 63.37 & +0.90 & 0.9023 & 0.9005 & -0.0018 & 0.8479 & 0.8449 & -0.0030 \\
Phi-4            & 42.58 & 44.35 & +1.77 & 59.63 & 61.37 & +1.74 & 0.8936 & 0.9000 & +0.0064 & 0.8334 & 0.8355 & +0.0021 \\
DeepSeek-R1 32B  & 38.87 & 40.94 & +2.07 & 56.25 & 58.26 & +2.01 & 0.8827 & 0.8816 & -0.0011 & 0.8194 & 0.8140 & -0.0054 \\
DeepSeek-R1 14B  & 38.43 & 33.33 & -5.10 & 55.13 & 51.14 & -3.99 & 0.8732 & 0.8506 & -0.0226 & 0.8031 & 0.7701 & -0.0330 \\
Llama3.1         & 35.08 & 29.37 & -5.71 & 52.10 & 48.57 & -3.53 & 0.8717 & 0.8573 & -0.0144 & 0.7925 & 0.7664 & -0.0261 \\
DeepSeek-R1 8B   & 30.89 & 30.26 & -0.63 & 48.33 & 48.05 & -0.28 & 0.8558 & 0.8331 & -0.0227 & 0.7632 & 0.7517 & -0.0115 \\
Mistral          & 26.62 & 27.79 & +1.17 & 44.89 & 45.00 & +0.11 & 0.8444 & 0.8490 & +0.0046 & 0.7436 & 0.7537 & +0.0101 \\
Llama3.2         & 25.19 & 22.38 & -2.81 & 42.47 & 43.10 & +0.63 & 0.8311 & 0.8190 & -0.0121 & 0.7297 & 0.7041 & -0.0256 \\
\hline
\end{tabular}}
\caption{Translation task on the \textbf{lexical choice benchmark} with a structured (XML-style) reasoning prompt.
Scores over 200 examples (validation+test): without reasoning (w/o), with reasoning (w/), and \(\Delta = \text{w/} - \text{w/o}\) (positive = gain, negative = drop).
The XML tags constrain both reasoning and the final translation, reducing format errors and easing parsing.}
\label{tab:lexical-choice-xml-reasoning}
\end{table}
}

\FloatBarrier

\begin{figure}[ht]
  \centering
  \includegraphics[width=.80\linewidth]{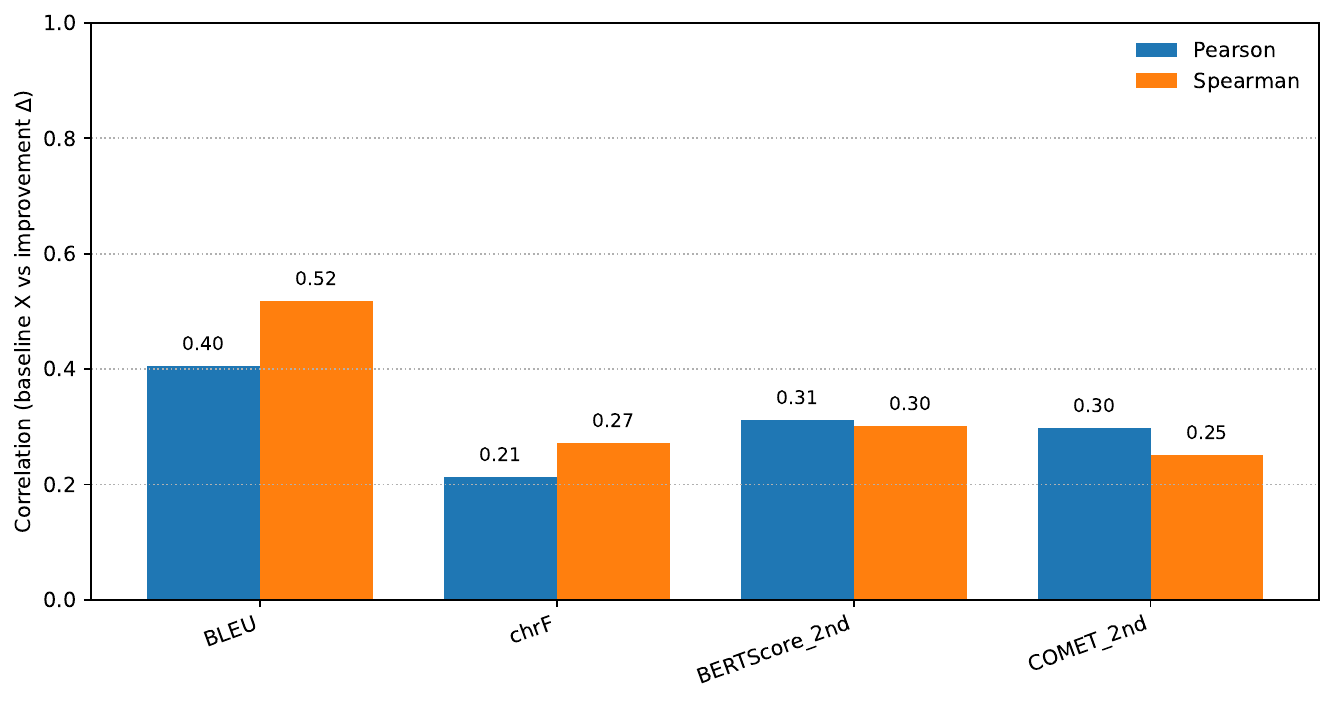}
  \caption{Correlation between baseline (no-reasoning) scores and gains \(\Delta\) under the structured (XML-style) reasoning prompt on the \textbf{lexical choice} task.
  Bars report Pearson and Spearman coefficients for BLEU, chrF, BERTScore, and COMET.
  Positive values across metrics support the “wise get wiser” effect: higher-scoring models tend to benefit more from reasoning.}
  \label{fig:app-gen-lex-wise-bar}
\end{figure}

\begin{figure}[ht]
  \centering
  \includegraphics[width=.99\linewidth]{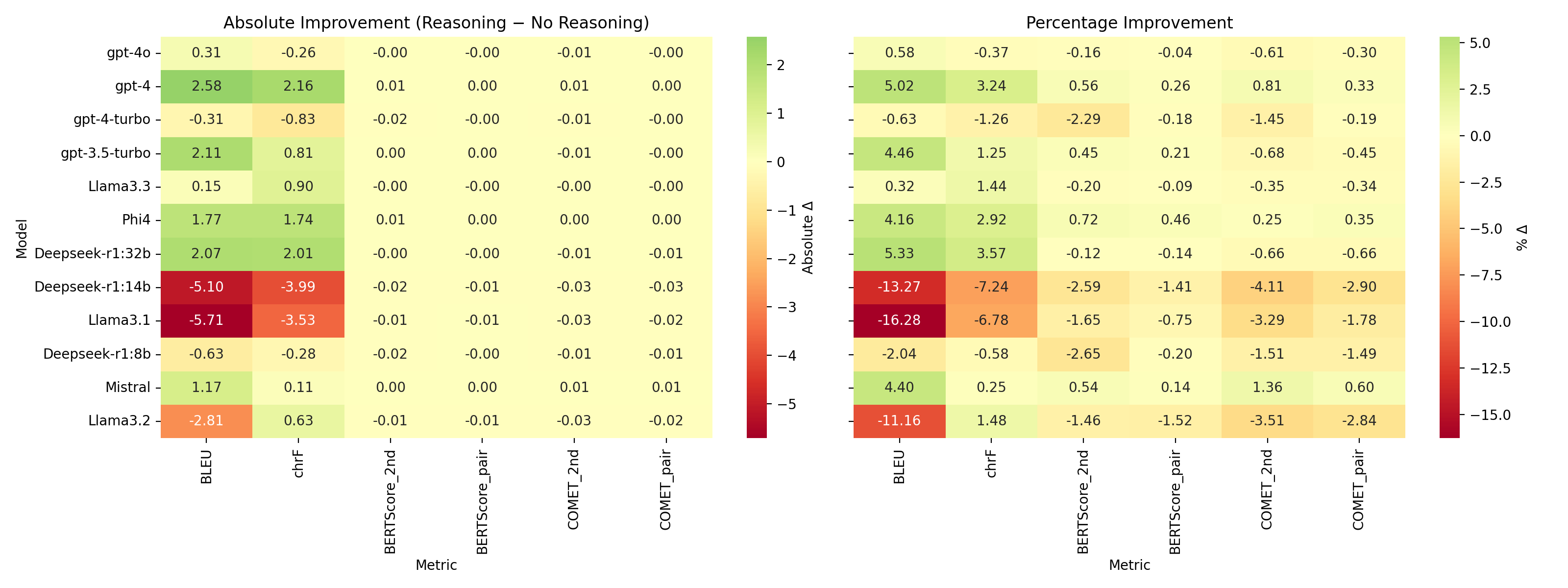}
  \caption{Heatmaps of gains from the structured (XML-style) reasoning prompt on the \textbf{lexical choice} task. 
  Left panel shows absolute changes for BLEU, chrF, BERTScore, and COMET; right panel shows percentage changes relative to the no-reasoning baseline. 
  Positive cells indicate improvements, negative cells declines, enabling a quick comparison across models and metrics.}
  \label{fig:app-gen-lex-improv-heatmaps}
\end{figure}

\begin{figure}[ht]
  \centering
  \includegraphics[width=.80\linewidth]{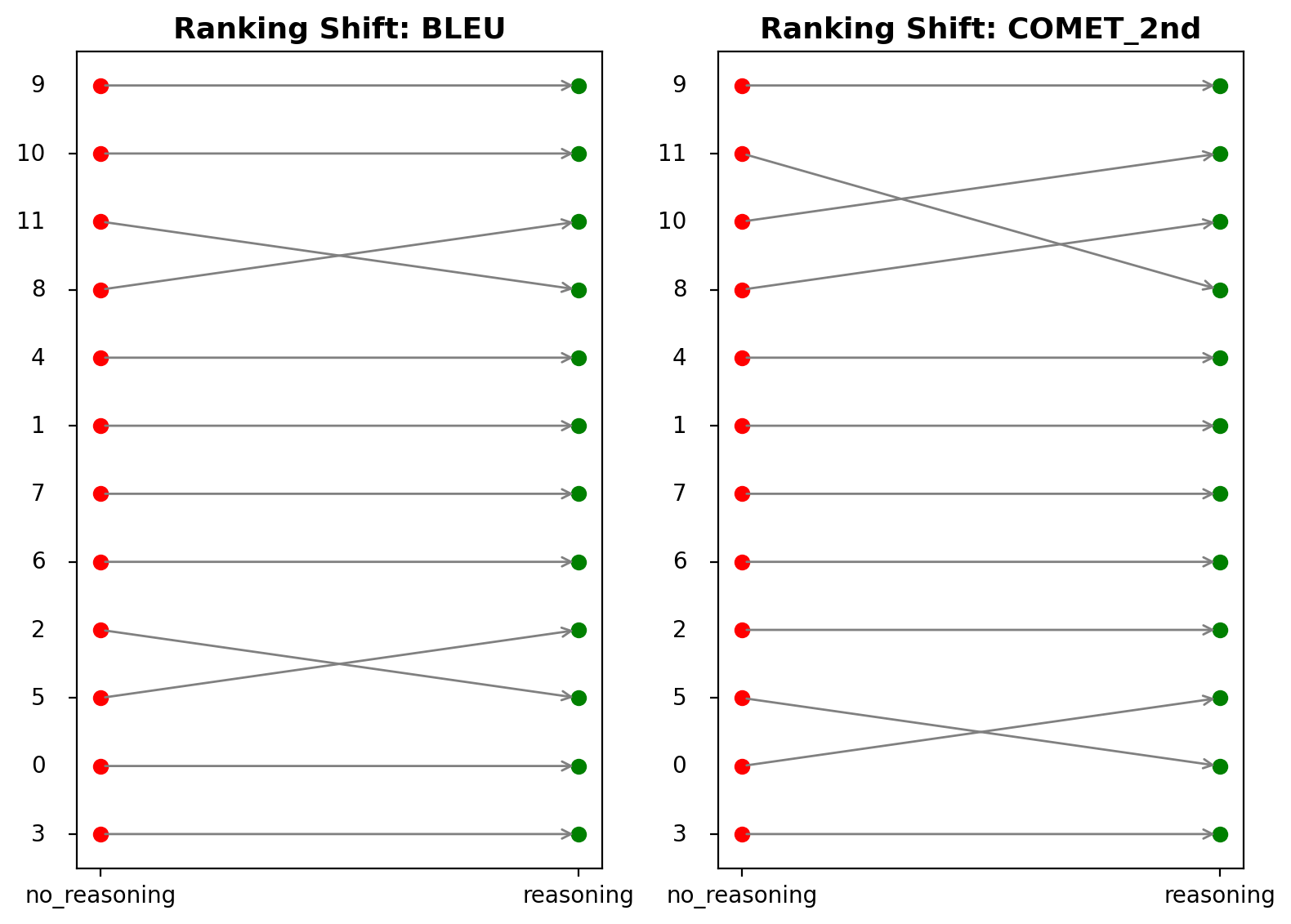}
  \caption{Ranking shifts from no\_reasoning (left) to structured reasoning (right) for BLEU and COMET.
  Lines show how each model’s relative position changes: upward trajectories signal gains; crossings indicate reorderings between models.
  Focus on these two metrics makes differences in lexical fidelity and semantic adequacy easy to compare.}
  \label{fig:app-gen-lex-ranking-shifts-blue-comet}
\end{figure}

\begin{figure}[ht]
  \centering
  \includegraphics[width=.80\linewidth]{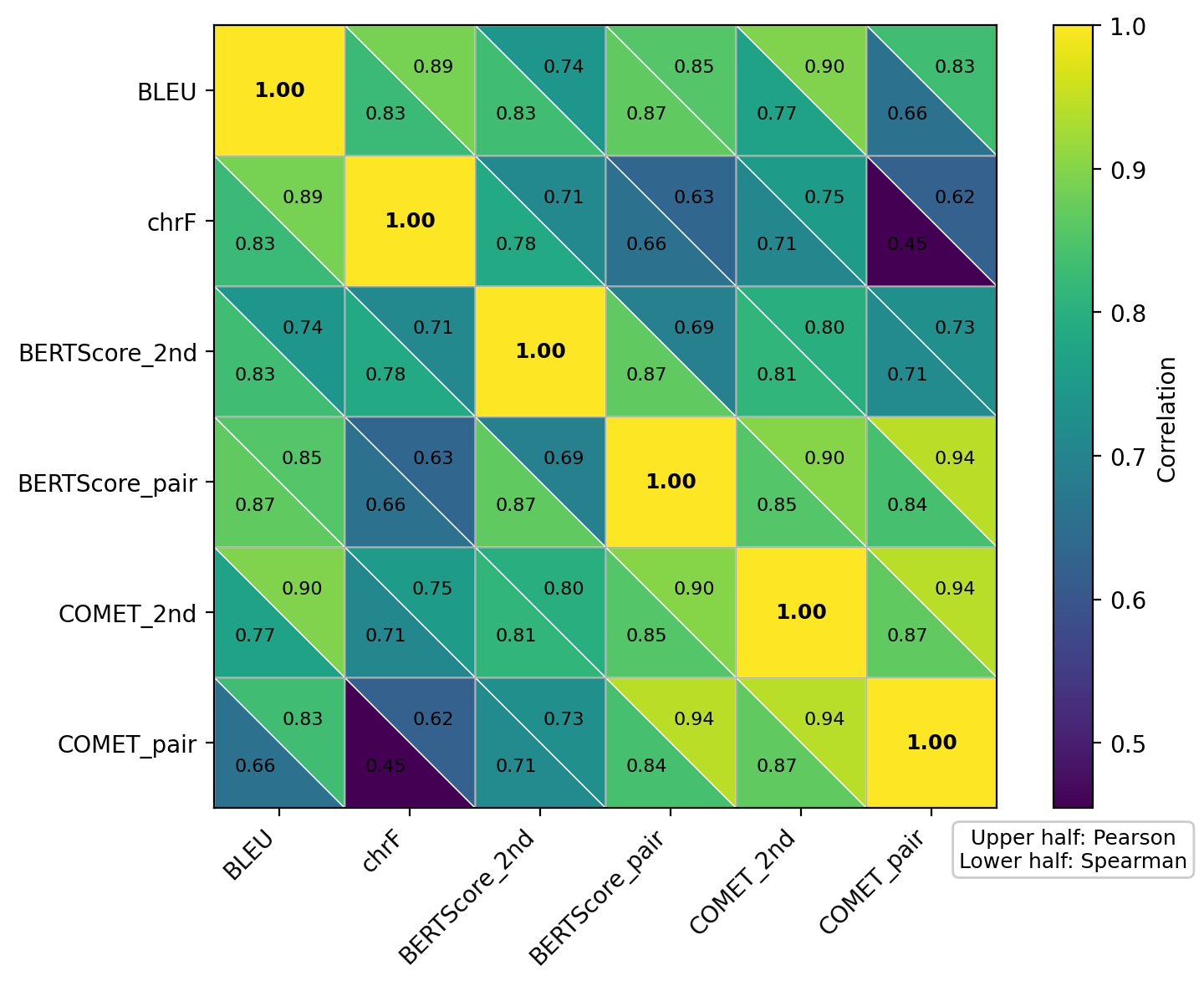}
  \caption{Correlation matrix of improvements \(\Delta\) (reasoning \( - \) no\_reasoning) under the structured (XML-style) prompt on the \textbf{lexical choice} task.
  Each cell reports Pearson (upper triangle) and Spearman (lower triangle) correlations across models.
  Positive cells indicate that gains tend to move together across metrics, while negative cells suggest trade-offs.}
  \label{fig:app-gen-lex-corr}
\end{figure}

\end{document}